\documentclass[11pt]{article}
\usepackage[]{emnlp2021}
\usepackage{times}
\usepackage{latexsym}
\usepackage{graphicx}
\usepackage{shortcuts}
\usepackage{amssymb}
\usepackage{amsmath}
\usepackage{booktabs}
\usepackage{url} 
\usepackage{enumitem}
\usepackage{mathtools}
\usepackage{subcaption}
\usepackage{fontawesome} 
\usepackage{multirow}
\usepackage{contour}

\usepackage{microtype}

\newcommand{\para}[1]{\smallskip\noindent\textbf{#1}}

\begin{document}

\title{RICA: Evaluating Robust Inference Capabilities \\Based on Commonsense Axioms}

\author{
Pei Zhou \quad Rahul Khanna \quad Seyeon Lee \quad Bill Yuchen Lin
\quad Daniel Ho \\ \textbf{Jay Pujara \quad Xiang Ren}\\
Department of Computer Science and Information Sciences Institute\\
University of Southern California\\
\small{\texttt{\{peiz,rahulkha,seyeonle,yuchen.lin,hsiaotuh,jpujara,xiangren\}@usc.edu}}
}

\maketitle

\newcommand{\rica}{\textsc{RICA}}
\newcommand{\earg}{entity argument}
\newcommand{\sarg}{semantic argument}

\begin{abstract}

Pre-trained language models (PTLMs) have achieved impressive performance on commonsense inference benchmarks, but their ability to employ commonsense to make robust inferences, which is crucial for effective communications with humans, is debated.
In the pursuit of advancing fluid human-AI communication, we propose a new challenge, \rica: \textbf{R}obust \textbf{I}nference using \textbf{C}ommonsense \textbf{A}xioms, that evaluates robust commonsense inference despite textual perturbations. To generate data for this challenge, we develop a systematic and scalable procedure using commonsense knowledge bases and probe PTLMs across two different evaluation settings. Extensive experiments on our generated probe sets with more than 10k statements show that PTLMs perform no better than random guessing on the zero-shot setting, are heavily impacted by statistical biases, and are not robust to perturbation attacks. We also find that fine-tuning on similar statements offer limited gains, as PTLMs still fail to generalize to unseen inferences. Our new large-scale benchmark exposes a significant gap between PTLMs and human-level language understanding and offers a new challenge for PTLMs to demonstrate commonsense.\footnote{Our code and data are public at \url{https://sites.google.com/usc.edu/rica}.}

\commentout{
(TO BE UPDATED) Pre-trained language models (PTLM) have greatly improved performance on commonsense inference benchmarks, but it remains unclear whether they are  pattern matching or reasoning like humans.
Although prior studies have found success in PTLMs' abilities to recall factual knowledge and reason with explicitly stated rules, a key property of reasoning is unexplored: \emph{robust axiomatic logical inference}.
Since humans can make logical inferences based on commonsense and are tolerant of language variations, this capability is essential for models to reach human-level intelligence.
We thus propose a challenge that evaluates robust axiomatic deductive inference in PTLMs in three task settings and develop a systematic procedure to generate probes that target these capabilities.
Extensive experiments on our generated probe sets show that PTLMs perform no better than random guessing on our probes(\textit{even with fine-tuning}), are heavily dependent on statistical biases, and are not robust to perturbation attacks.
Our framework and initial probe sets can help future work in improving PTLMs' logical inference abilities and robustness to linguistic variations.\footnote{\small Our data and code are submitted as supplementary material and will be public later.}
}
\end{abstract}

\section{Introduction}\label{intro}
\begin{figure}[t]
    \hspace{-0.3cm}
	\includegraphics[width=1.02\columnwidth]{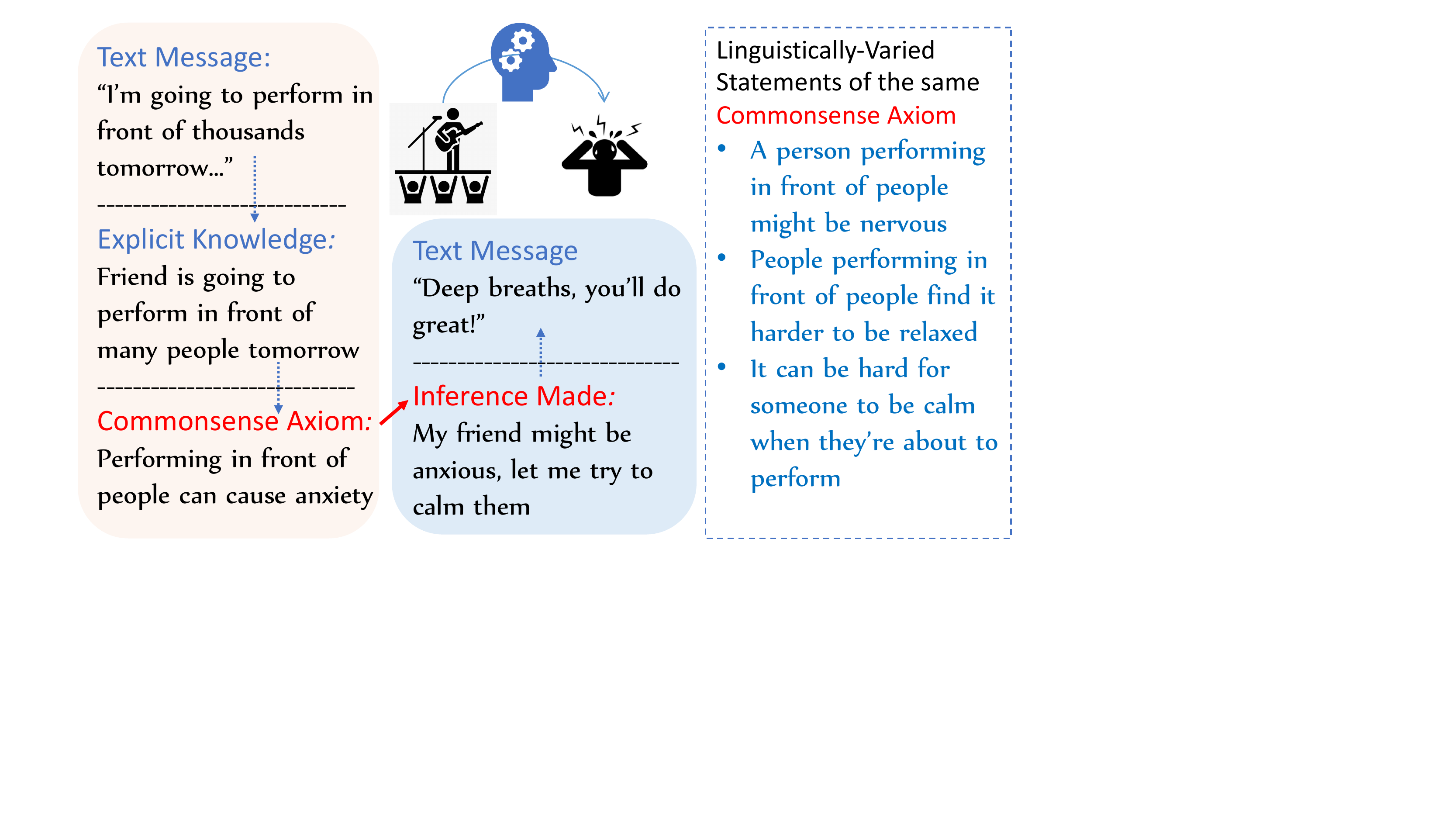}
	\caption{
	{Human communication requires commonsense inferences. \rica~evaluates such inferences via commonsense axioms with many linguistic variations.}}
	\label{fig:motivation}
\end{figure}


Smooth and effective communication requires the ability to make various forms of commonsense inferences~\cite{clark1991grounding}.
When a friend texts,  ``\textit{I'm going to perform in front of thousands tomorrow,}'' you may reply reassuringly, ``\textit{Deep breaths, you'll do great!}'' Implicit to this communication is a commonsense logical inference that a person performing in front of a crowd may feel anxious, and that a reassuring remark helps ease anxiety (Figure~\ref{fig:motivation}). 
A growing body of literature~\cite{bosselut2019comet, petroni2019language} shows pre-trained language models (PTLMs) are able to catalog the types of commonsense relationships necessary for fluid communication. However, as we show in this paper, PTLMs have a shocking inability to leverage such commonsense knowledge to make robust inferences. 

Here we focus on two specific characteristics crucial to human-AI communications: (1) combining commonsense knowledge with information expressed in natural language to make inferences and (2) producing consistent inferences amidst logically-equivalent yet linguistically-varied paraphrases. We focus on \emph{commonsense axioms}, such as ``\textit{Performing in front of people can cause anxiety}'', and exploit the flexibility of language to express the same axiom in many forms --- \textit{e.g.}, ``\textit{Performing in front of people makes it hard to stay calm.}" We test these characteristics by generating self-contained commonsense statements involving novel entities (``\textit{Prindag is going to perform in front of a crowd, so prindag is more likely to feel nervous.}'') and adapt them to two evaluation settings.



\commentout{
likely to conclude that the person is feeling nervous, even though it has not been stated. 
Similar to this scenario, most interactions with humans require an assumed understanding of a certain \textit{commonsense axiom}. 
More formally, a \textit{commonsense axiom} is any relation pertaining to commonsense knowledge that is often assumed rather than stated in natural language. 
Moreover, as shown in Figure 1, in order to interact with humans in everyday scenarios, knowing the commonsense axioms is not enough and studies have shown evidence of this ability in some models~\cite{bosselut2019comet, petroni2019language}, but rather the AI agent must also be able to use the axioms in a chain of reasoning. 
Effective communication also requires the \textit{robustness to linguistic variations}, as each condition of an axiom might be stated differently, but the relationship of the axiom remains the same. 
It is thus imperative for AI agents to have the capabilities of \textit{reasoning with commonsense axioms} and being \textit{robust to language variations} in order to interact with humans~\cite{Mccarthy1960ProgramsWC,davis2015commonsense}.
}
Unfortunately, these two capabilities have largely been overlooked by existing natural language inference (NLI) benchmarks~\cite{MNLI} and knowledge probing studies for transformer-based PTLMs ~\cite{vaswani2017attention, Devlin2019, liu2019roberta,clark2020transformers, petroni2019language}.
Most existing commonsense reasoning-focused datasets~\cite{zhang2017ordinal,MNLI,ostermann-etal-2019-commonsense,zhou-etal-2021-commonsense, talmor2019commonsenseqa} do not systematically evaluate robustness against linguistic variations, meaning we cannot preclude the possibility that models are learning spurious patterns to solve the needed task. 

To fill this gap, we introduce \rica, a challenge to evaluate a model's \textbf{R}obust \textbf{I}nference using \textbf{C}ommonsense \textbf{A}xioms in English. 
\rica~draws on linguistic and cognitive science research~\cite{schank1977scripts, alshawi1989logical} suggesting humans translate language to logical representations and reason using these abstract representations. 
\rica~consists of a set of natural language statements in the ``premise-conclusion'' format that require reasoning using latent (implicit) commonsense relationships. 
We formulate these abstract commonsense relations between entities in first-order logic and refer to them as \emph{commonsense axioms} (see Fig.~\ref{fig:motivation}). To insulate from PTLM biases and test human-like acquisition ability on new words~\cite{carey1978acquiring},
\rica~uses \emph{novel entities}, which are unseen strings used to ground axioms into natural language.  
Finally, we introduce a set of \emph{linguistic perturbations} that paraphrase a commonsense axiom into natural language in various forms.

Each component of \rica~is generalizable, providing a systematic procedure to generate myriad commonsense statements. In this paper, we generate 257k commonsense statements capturing 43k axioms comprising different types of commonsense, such as physical, material, and social properties. To demonstrate the quality of \rica, we create a manually-curated set of 1.6k probes based on commonsense axioms, and also undertake a large-scale, crowdsourced verification of 10k generated statements with multiple human annotators. \rica~is built by leveraging existing commonsense knowledge bases such as ConceptNet~\cite{liu2004conceptnet} and ATOMIC~\cite{sap2019atomic} to support easy expansion. Furthermore, \rica's statements can be posed as popular PTLM tasks such as  masked word prediction or sentence probability, making our benchmark widely applicable.
\rica~provides an extensible platform for evaluating commonsense reasoning in a variety of PTLMs.

\begin{figure}[t]
	\centering
	\includegraphics[width=1.0\columnwidth]{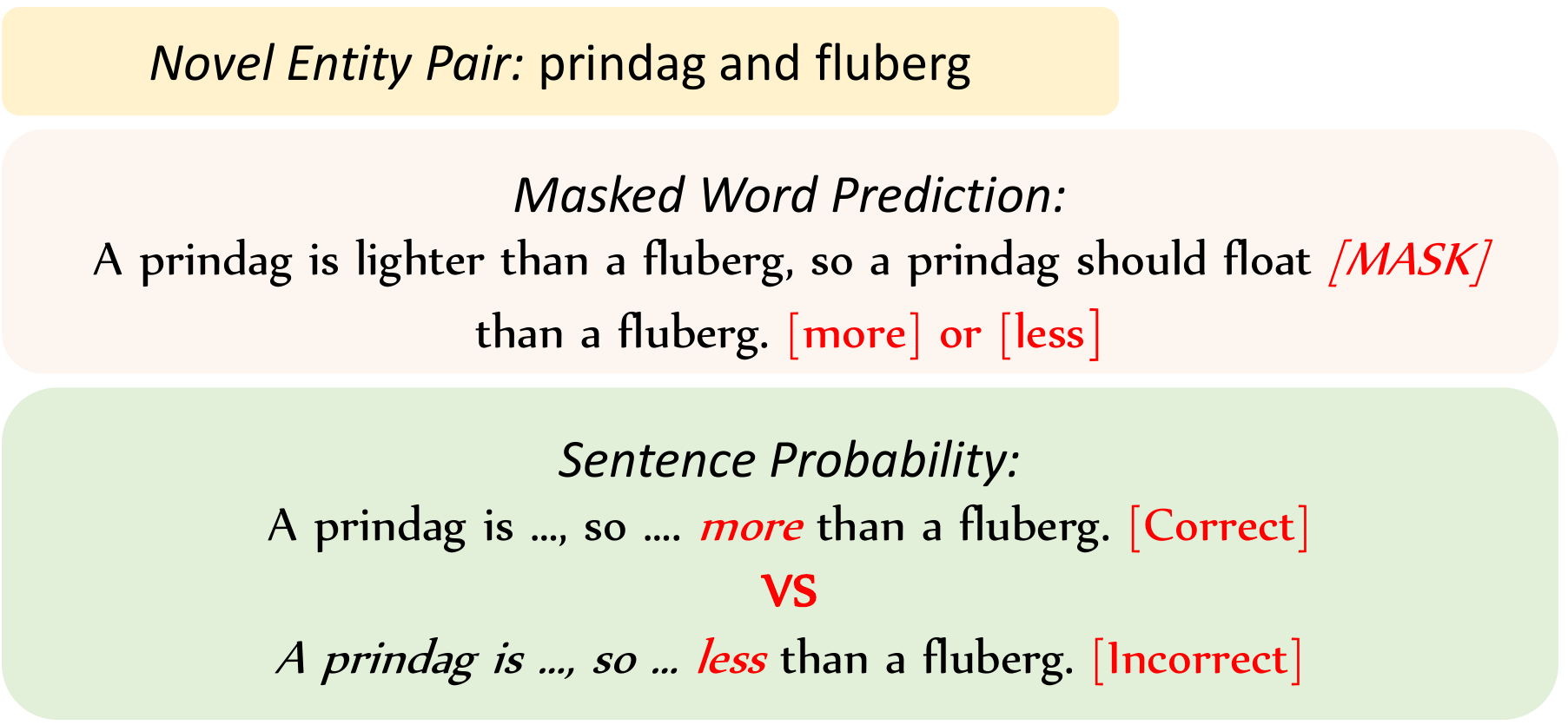}
	\caption{Illustration of two evaluation settings with a pair of novel entities used by \rica~probes.}
	\label{fig:eval_settings}
\end{figure}

When evaluating state-of-the-art transformer-based PTLMs on the \rica~probes following a \textit{zero-shot} setting (\textit{e.g.}, predicting ``\textit{more}" vs. ``\textit{less}" in the first example in Fig.~\ref{fig:eval_settings}), we consistently discover their performance is on par with \textit{random guessing}. 
Even after fine-tuning with large amounts of labeled examples, PTLMs exhibit a significant gap relative to human performance. 
We drill down into this finding through (1) zero-shot, (2) low-resource, (3) high-resource, and (4) noisy training settings and find that even with appreciable performance gains on automatically generated probes in high resource settings, PTLMs still remain on par with random guessing on difficult, human-curated \rica~probes. 
To better understand these results, we identify a pervasive intrinsic bias in PTLMs that demonstrates positivity bias in human languages~\cite{dodds2015human}.
\section{The RICA Challenge}\label{procedure}
\begin{figure*}[t]
	\centering
	\includegraphics[width=0.92\linewidth]{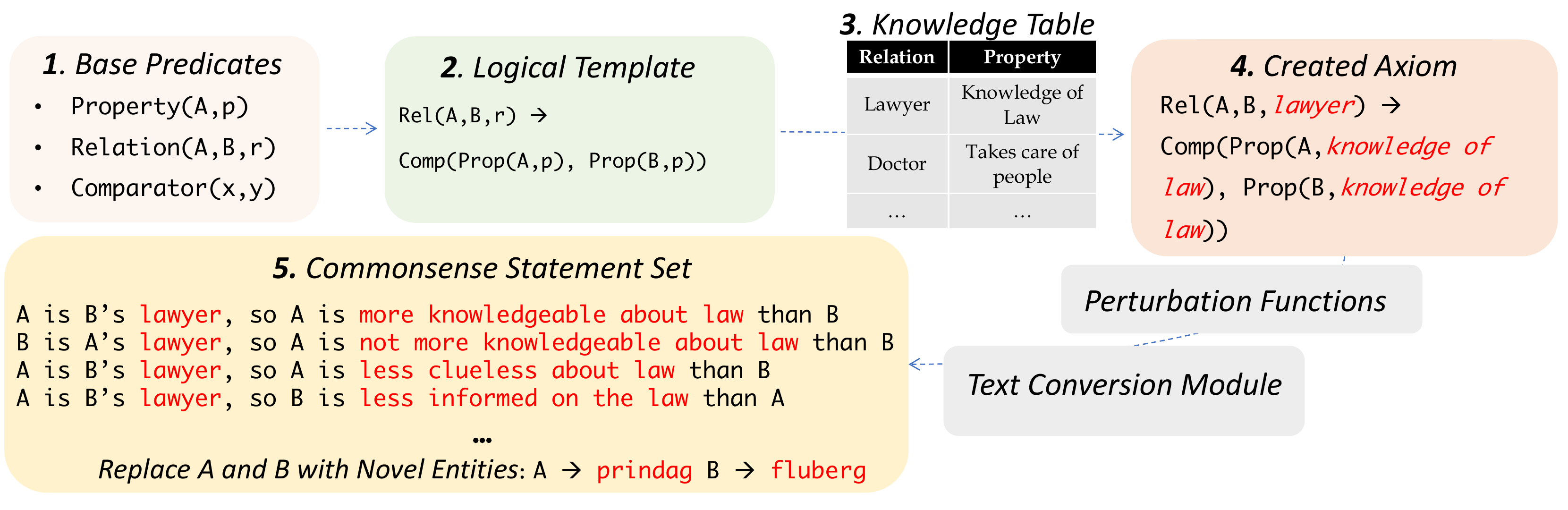}
	\caption{\textbf{Overview of the workflow of our statement construction process.} The output is a set linguistically-diverse of masked sentences that follow the same reasoning template.}
	\label{fig:pipeline}
\end{figure*}

The \rica~challenge is posed as a set of textual statements (sentences), each expressing a latent commonsense relationship in the ``premise-conclusion'' format (see Stage 5 in Fig.~\ref{fig:pipeline} for examples).
These statements use generated novel entities such as ``\textit{prindag}'' and ``\textit{fluberg}'' instead of real-world entities such as ``\textit{thimble}'' and ``\textit{elephant}'' to separate factual recalling from reasoning. 
Each statement can be viewed as an instantiation of a commonsense principle, such as ``\textit{smaller objects cannot contain larger objects.}'' 

We express these commonsense principles in first-order logic, further generalizing statements through the use of general predicates for object properties (\textit{e.g.}, size) and object-object relations (\textit{e.g.}, containment).
We turn these logical formulae into the associated textual statements using a set of perturbation operators and a conversion module, which together produce a logically-equivalent set of commonsense statements. 
In the rest of this section, we first provide a formal definition of RICA challenge, then provide a detailed description of the statement construction process.


\subsection{Challenge Formulation}
\label{sec:challange-formulation}
Formally, we define a commonsense axiom $a_i$, expressed via a first-order-logic (FOL) formula, as a relationship between entities that can be inferred using commonsense knowledge (see Stage 4 in Fig.~\ref{fig:pipeline}).
To test whether PTLMs understand an axiom $a_i$, as well as examine their robustness to linguistic variations, we instantiate the axiom $a_i$ by a set of $m$ syntactically-different commonsense statements $\left \{ s^i_{1}, s^i_{2}, ..., s^i_{m} \right \}$, each expressing the \textit{same} logic as the axiom.
Each statement takes the form of an inferential implication with a premise and conclusion. 
Finally, depending on the PLTM, we select an appropriate task (Section~\ref{setup}), transform each statement in the set into its task-specific \textit{probe}, and evaluate how well the PTLM can leverage the logic of $a_i$ to solve each of $a_i$'s corresponding probes.
We deem a model ``successful'' on the challenge (or, understands the axioms) only if it can perform like humans on all probes of the axioms.

\begin{table}[t]
\centering
\scalebox{0.63}{
\begin{tabular}{c|l}
\toprule
\multicolumn{1}{c}{\textbf{\textsc{Terminology}}} & \multicolumn{1}{|c}{\textbf{\text{Description}}}                                                                                       \\ \midrule
\textit{\textbf{Logical Template}} (LT)           & \begin{tabular}[c]{@{}l@{}}General FOL formula constructed from \\ predicates and logical connectives\end{tabular}    \\ \hline
\textit{\textbf{Arguments}}                       & \begin{tabular}[c]{@{}l@{}}Specific entities and relations to fill \\ predicates in LTs\end{tabular}                  \\\hline
\textit{\textbf{Axiom}}                           & \begin{tabular}[c]{@{}l@{}}Commonsense relationship expressed \\ in FOL by filling a LT with arguments\end{tabular}   \\\hline
\textit{\textbf{Commonsense Statement}}           & \begin{tabular}[c]{@{}l@{}}Natural language sentence after \\ converting an axiom using a TT\end{tabular}             \\\hline
\textit{\textbf{Statement Set}}                   & \begin{tabular}[c]{@{}l@{}}Statements that inform the same \\ axiom after applying perturbations\end{tabular}         \\\hline
\textit{\textbf{Evaluation Instances/Probe}}      & \begin{tabular}[c]{@{}l@{}}A set of statements after adapting \\ to an evaluation task\end{tabular}     
\\ \bottomrule
\end{tabular}
}
\caption{Description of terminology used in \rica.}
\label{tab:terminology}
\end{table}

\commentout{
If not here, maybe in the appendix have the algorithm:
for LT,TT: {P, M, S, T}
    for args: LT
        gargs = ground(args, kt)
        sent = text(gargs, TT)
        for p : perturbations:
            stmt += p(sent)
        stmt+= sent
Obviously clean this up and match the formalization in 2.1
}


\subsection{Statement Set Construction Process}\label{subsec:fol}
This subsection introduces our proposed procedure for the construction of commonsense inference statement sets for the challenge.
A list of terminologies and descriptions can be found in Table~\ref{tab:terminology} and an overview of our workflow is shown in Figure~\ref{fig:pipeline}.

\para{Stage 1. Define Predicates.}
In FOL, predicates are used to denote a property of objects or a relation between objects and every predicate symbol comes with an arity larger or equal to 1.
We define three general high-level predicates that serve as the backbone for the logical formulations of our axioms: \textit{Property}, \textit{Comparator} and \textit{Relation}.
(1) $\textbf{\textsc{Prop}}(A, p)$ represents that entity $A$ has a certain property $p$. ``$\textsc{Prop}(A, glass)$'' indicates that A is made of glass.
(2) $\textbf{\textsc{Rel}}(A, B, r)$ represents that $A$ and $B$ have a certain relation $r$. ``$\textsc{Rel}(A, B, lawyer)$'' indicates that A is B's lawyer.
(3) $\textbf{\textsc{Comp}}(x,y)$ represents a comparative relationship between values $x$ and $y$, where ``$\textsc{Comp}$'' will be replaced with comparison words like ``better," ``more," or ``easier."
We will later define multiple sub-types of these predicates to crawl from Knowledge Bases (KBs) to ensure a wide coverage of common knowledge.

\para{Stage 2. Compose Logical Templates.}
We manually create first-order logical formulae, referred to as \textit{logical templates} (LT), using the predicates defined in Stage 1. Each formula takes the form of an implication, expressing an inference based on commonsense knowledge. For example, \begin{small}$\textsc{Rel}(A,B,r) \rightarrow  \textsc{Comp}(\textsc{Prop}(A,p), \textsc{Prop}(B,p))$\end{small} expresses a logical inference comparing a property of two entities, $A$ and $B$, based on a relationship between them.
An instantiated version of this template can be
\begin{small}$\textsc{Rel}(A,B,lawyer) \rightarrow  \textsc{More}(\textsc{Prop}(A, know\_law), \textsc{Prop}(B,know\_law))$\end{small}.

\para{Stage 3. Populating Knowledge Tables.}
Materializing the abstract relationships in a logical template requires connecting abstract logic to commonsense knowledge. 
We define a structure called \textit{knowledge table} (KT) that contains valid arguments to populate a specific LT and form a FOL representation of the axiom. KTs are generated by crawlining commonsense KBs such as ConceptNet~\cite{liu2004conceptnet} and ATOMIC~\cite{sap2019atomic}. 
The first step of the crawling process is to narrow down the predicates to specific types. For example, \textsc{Prop} is general enough to capture an entity's capabilities (e.g., \textit{knowledge of law}) or its intrinsic properties (e.g., \textit{hardness}).
We pre-define several type constraints for both properties (\textsc{Prop}) and relations (\textsc{Rel}). For \textsc{Prop}, we consider \emph{Capability, Attribute, and Condition}. For \textsc{Rel}, we consider \emph{Role and Action}. Note that these categories can be extended for wider coverage of knowledge and allow our LTs to be adapted to a broader range of KB schemas.
After specifying type constraints, we specify steps for crawling the arguments either from commonsense KBs such as Concept and ATOMIC or general web KB such as Wikipedia. In our example in Fig.~\ref{fig:pipeline}, we can crawl occupations from Wikipedia, and then query ConceptNet for triples with the occupation as the subject and \emph{CapableOf} as the relationship to create a KT with professions and capabilities. We show all crawling steps for KTs in Appendix~\ref{data_details}.

\para{Stage 4. Creating Axioms.}
Combining knowledge tables and logical templates allows us to generate commonsense axioms at scale, which are partially-filled LT formulae.
For example in Fig.~\ref{fig:pipeline} Stage 3, the arguments of predicates \textsc{Rel}, \textsc{Prop}, and \textsc{Comp} are set in order to reflect the commonsense relationship between \textit{lawyer} and \textit{knowledge of law}, while leaving the entities A and B ungrounded. 
Once the predicates are instantiated, we call this partially-filled LT a \emph{commonsense axiom}.

    



\para{Stage 5. Generate Statement Sets.}
After filling the logical templates, each partially-filled LT represents one commonsense axiom. 
To comprehensively challenge models' understanding of an abstract axiom, we construct a \emph{statement set} expressing the \textit{same axiom} with \textit{different phrasings}, i.e., logically-equivalent yet linguistically-varied.
We define several \textit{perturbations} to apply on the \textit{arguments} from knowledge tables.

\begin{table}[tb]
\centering
	\scalebox{0.65}{
\begin{tabular}{@{}c c@{}}
\toprule
\textsc{Linguistic Operator}  & \textsc{Example}                                                                                                               \\ \midrule
\textsc{Negation}                                              & \textsc{Neg}(fit into) = not fit into                                                                                                   \\ 
\textsc{Antonym}                                                 & \textsc{Ant}(fit into) = contain                                                                                                        \\ 
\textsc{Paraphrase}                                                & \textsc{Para}(fit into) = put into                                                                                                      \\ \midrule
\textsc{Paraphrase Inversion}                                 & \begin{tabular}[c]{@{}c@{}}\textsc{Para}(\textsc{Ant}(fit into)) = Para(contain)\\ = hold inside\end{tabular}                                                                             \\ 
\textsc{Negation Antonym}                               & \begin{tabular}[c]{@{}c@{}}\textsc{Neg}(\textsc{Ant}(fit into)) = \textsc{Neg}(contain)\\ = not contain\end{tabular}                                                                                 \\ 
\textsc{Negation Paraphrase}                                  & \begin{tabular}[c]{@{}c@{}}\textsc{Neg}(\textsc{Para}(fit into)) = \textsc{Neg}(put into)\\ = not put into\end{tabular}                                                                              \\ 
\textsc{Negation Para\_Inv}                    & \begin{tabular}[c]{@{}c@{}} \textsc{Neg}(\textsc{Para}(\textsc{Ant}(fit into))) = \textsc{Neg}(\textsc{para}(\\ contain))= \textsc{Neg}(hold inside)\\  =not hold inside \end{tabular} \\ \bottomrule
\end{tabular}
}
\caption{Linguistic operators, logic, and examples.}
\label{tab:LingOperators}
\end{table}
(1) \textit{Linguistic Operators.}
We define seven types of linguistic operators to facilitate and formalize perturbations, shown in Table~\ref{tab:LingOperators}.
We construct the last four operators by combining some of the single operators listed in the first three rows.
Note that for \textsc{Negation}, \textsc{Antonym}, \textsc{Paraphrase Inversion}, and \textsc{Negation Paraphrase} types, the logic of the original phrase is changed, so words in the statements have to be changed accordingly. For example, if we apply \textsc{Antonym} to ``\textit{fit into}'' in the probe ``\textit{A is smaller than B, so A is more likely to fit into B},'' we will get ``\textit{A is smaller than B, so A is \textbf{less} likely to \textbf{contain} B.''}
(2) \textit{Asymmetry Operator.}
Most of our logical templates use several strongly-ordered comparisons and relationships allowing us to introduce asymmetries that preserve meaning.
For example, $\textsc{More}(A,B) \rightarrow \neg \textsc{More}(B,A)$ and $\textsc{Rel}(A,B,parent) \rightarrow \neg \textsc{Rel}(B,A,parent)$. 
Using this invariant, we can swap the positions of two entities for these predicates and the logic will also be negated, so we denote this perturbation as $\textsc{Asym}(\textsc{P}(A,B)) \rightarrow \textsc{P}(B,A) = \neg \textsc{P}(A,B)$.

We apply the defined operators to the arguments in the predicates to first form a set of partially-filled LTs (axioms) and use for a conversion module to convert axioms to statements with diverse perturbations. In practice, this module can be a sequence-to-sequence (seq2seq) model (that takes in FOL and outputs natural language text), or human-written templates.
Finally, commonsense axioms are general logical relationships that hold for all entities. To formulate specific commonsense statements, we generate specific \emph{novel entities}. These entities are randomly generated character strings from length 3 to 12 that are not seen in the training data of the PTLMs. 
Using novel entities enables us to avoid conflating fact-based recall with commonsense reasoning when evaluating PTLMs.

\section{Experiment Setup}\label{setup}

\begin{figure}[t]
	\centering
	\includegraphics[width=0.85\columnwidth]{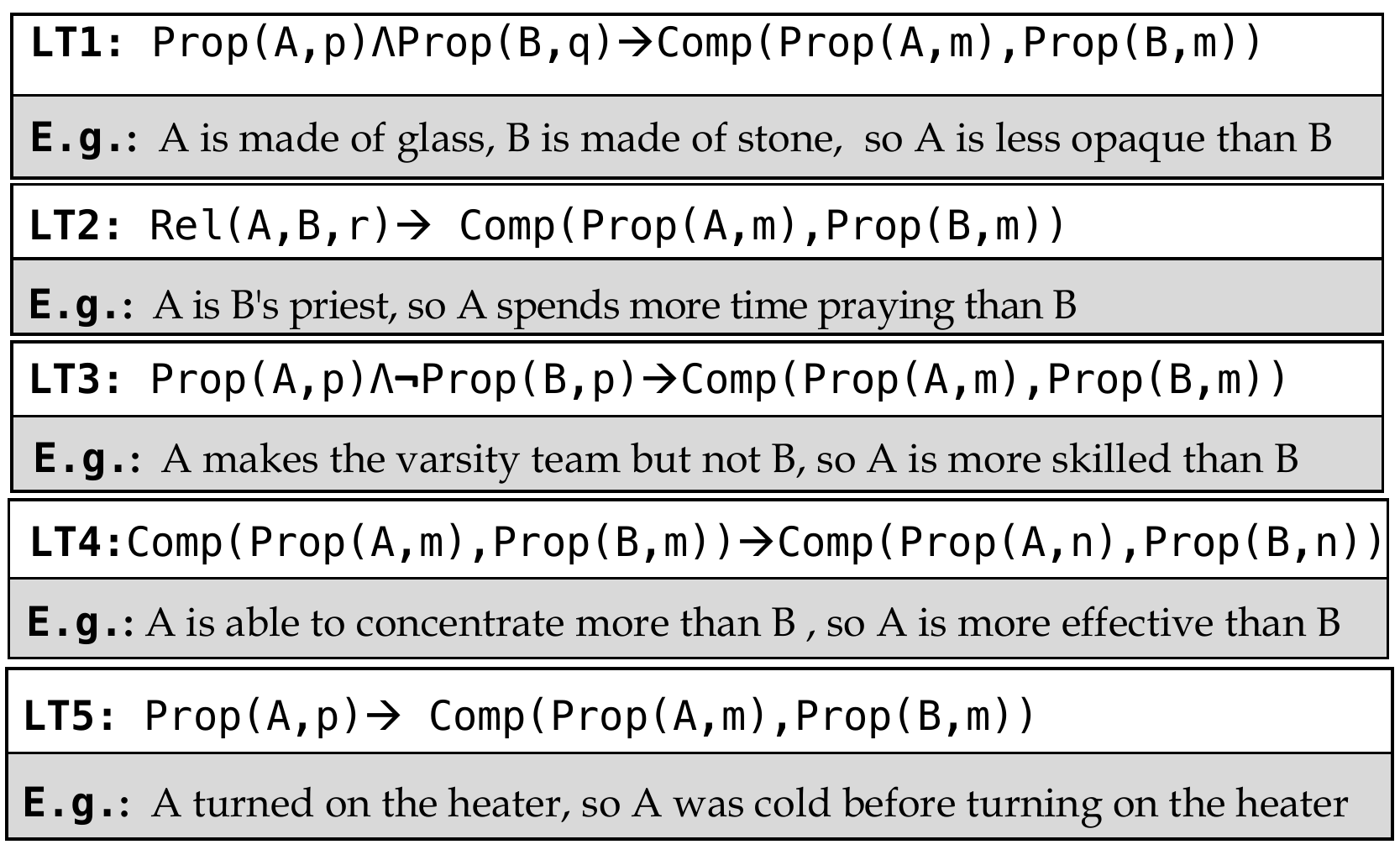}
	\captionof{table}{Example first-order logical templates we construct for our probes and an example for each template.}
    \label{tab:logical_templates}
\end{figure}

\subsection{Probing Tasks}
\label{formulation}
To examine transformer-based PTLMs' performance on RICA challenge, 
we draw conclusions from evaluation results on two distinct probing tasks shown in Figure~\ref{fig:eval_settings}, described as follows.

\para{Masked Word Prediction (MWP)~}
Inspired by the masked word prediction objective in BERT~\cite{Devlin2019}, we examine if the models can recover masked-out keywords in the statement given the remaining context. 
Since \rica's statements take the form of implications, we mask words in the consequent to evaluate the inference performance, given the premise. 
Specifically, we choose to mask the comparative words (from~\textsc{Comp}) such as `\textit{`more/less}'' and ``\textit{better/worse}'' as shown in Figure~\ref{fig:eval_settings}, since they not only capture the commonsense relationship, but also focus on masking positions where only a few options are appropriate logically and syntactically. 

\para{Sentence Probability (SP)} evaluates if PTLMs assign higher probability for statements that express commonsense axioms versus contradictory statements.
RICA statements are input to PTLMs, computing SP by taking the product of each word's probability conditioned on previous words, \textit{i.e.}, the left-to-right language modeling loss. 
For each RICA statement, we pair it with an incorrect (non-commonsense) statement by swapping the comparative word (\textit{i.e.}, the masked word in MWP) with its opposite word, as shown in Figure~\ref{fig:eval_settings}. 

\subsection{Probing Data Details}
\label{probing_data_details}
\para{Raw Set~}
Following the process in Section~\ref{procedure}, we use the three high-level predicates to generate five LTs as shown in \tabref{tab:logical_templates}. 
Then we construct knowledge tables to fill in each template by crawling from two commonsense KBs: ConceptNet~\cite{liu2004conceptnet} and ATOMIC~\cite{sap2019atomic}. 
Specifically, for each LT, we design 1 to 4 crawling strategies based on the type constraints we impose on the predicates so that it covers multiple aspects of commonsense knowledge (for all strategies please see Table~\ref{tab:strategies} in Appendix~\ref{data_details}). For example, the example shown for LT1 in \tabref{tab:logical_templates} is about inference of physical properties based on the material of two objects as we constrain $\textsc{PROP}$ in the premise to be materials. However, we can also constrain $\textsc{PROP}$ in the premise to be animals so that we can use the same template to examine inference of properties based on the animal types of A and B, e.g., ``A is a fish, B is a horse, so A is more likely to be in the bottom of the sea than B.''

We have 11 type-constrained LTs and we populate the KTs using 11 human-designed crawling strategies shown in Appendix~\ref{data_details}, resulting in around 43k axioms.
Then we apply the perturbation operators as described before to form a set of 257k perturbed axioms. 
For this large set, we apply negation and asymmetry operators automatically by adding negation and switching the order of entities.
To convert FOL axioms to text, we train a seq2seq model based on BART~\cite{lewis2019bart} on 200 manually converted axiom-text pairs covering each type-constrained LT and each perturbation type.
Finally, we replace entities to unseen entities to form a set of 257k commonsense statements.

\textit{Quality Check.}
To check for language quality of the generated probes from BART, we randomly sample 5\% of the 10k set and ask a native English speaker to check for the naturalness. We found that only \emph{4 out of 500} (0.8\%) probes contain grammar or fluency issues.
Since all probes follow a premise-conclusion format, we find that using 200 pairs of first-order logic (FOL) and aligned text for fine-tuning BART is sufficient to convert FOL into text, both from our manual inspection and the crowdsourcing verification of the generated probes. We tried increasing the training set size and didn’t observe a clear difference in quality.

    \begin{figure*}[tb]
    \centering
    \begin{subfigure}[t]{0.32\textwidth}
    \centering
        \includegraphics[width=\textwidth]{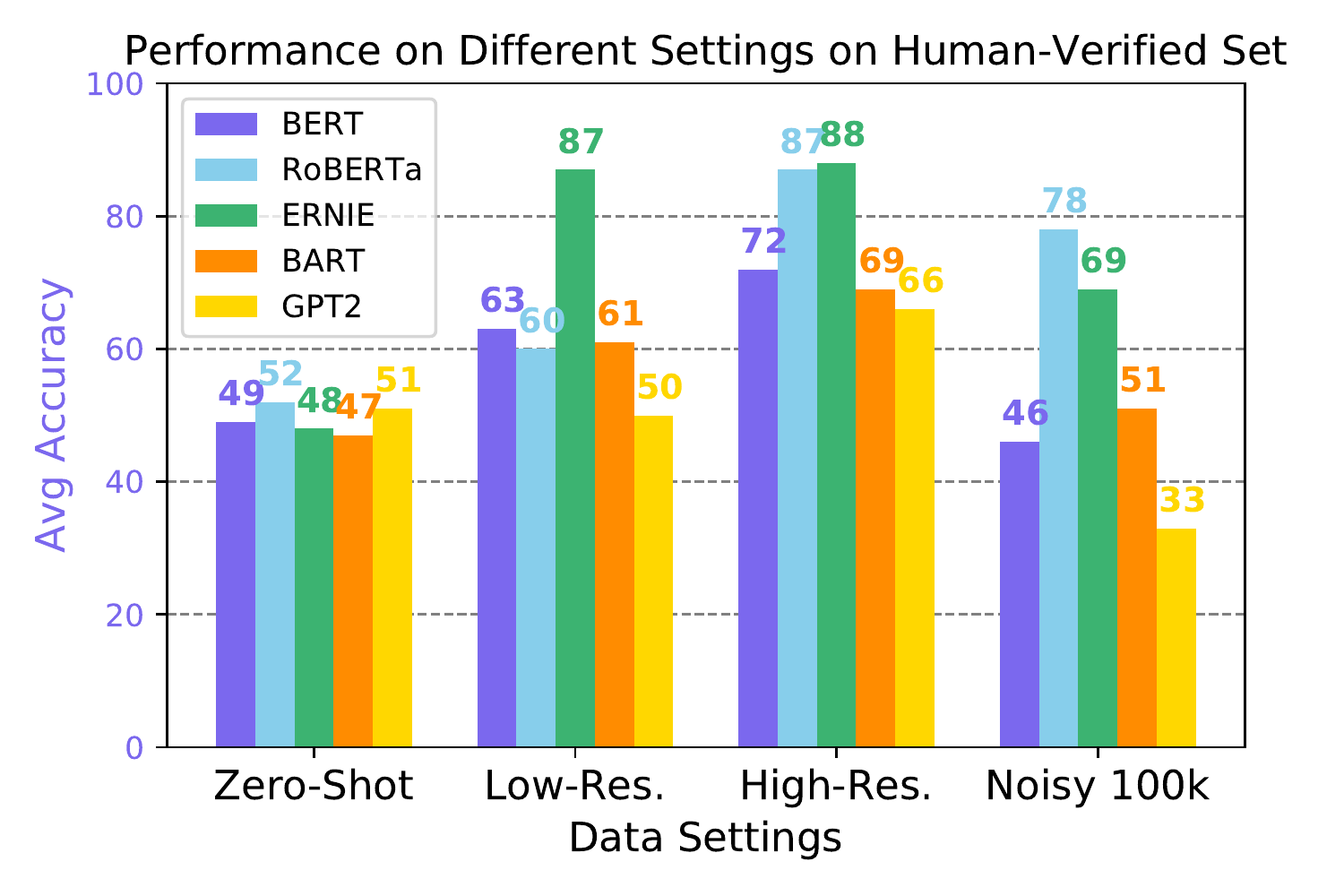}
        \caption{Performance on Human-Verified Set}
        \label{fig:avg_perf_easy}
    \end{subfigure}%
    \begin{subfigure}[t]{0.32\linewidth}
    \centering
        \includegraphics[width=\linewidth]{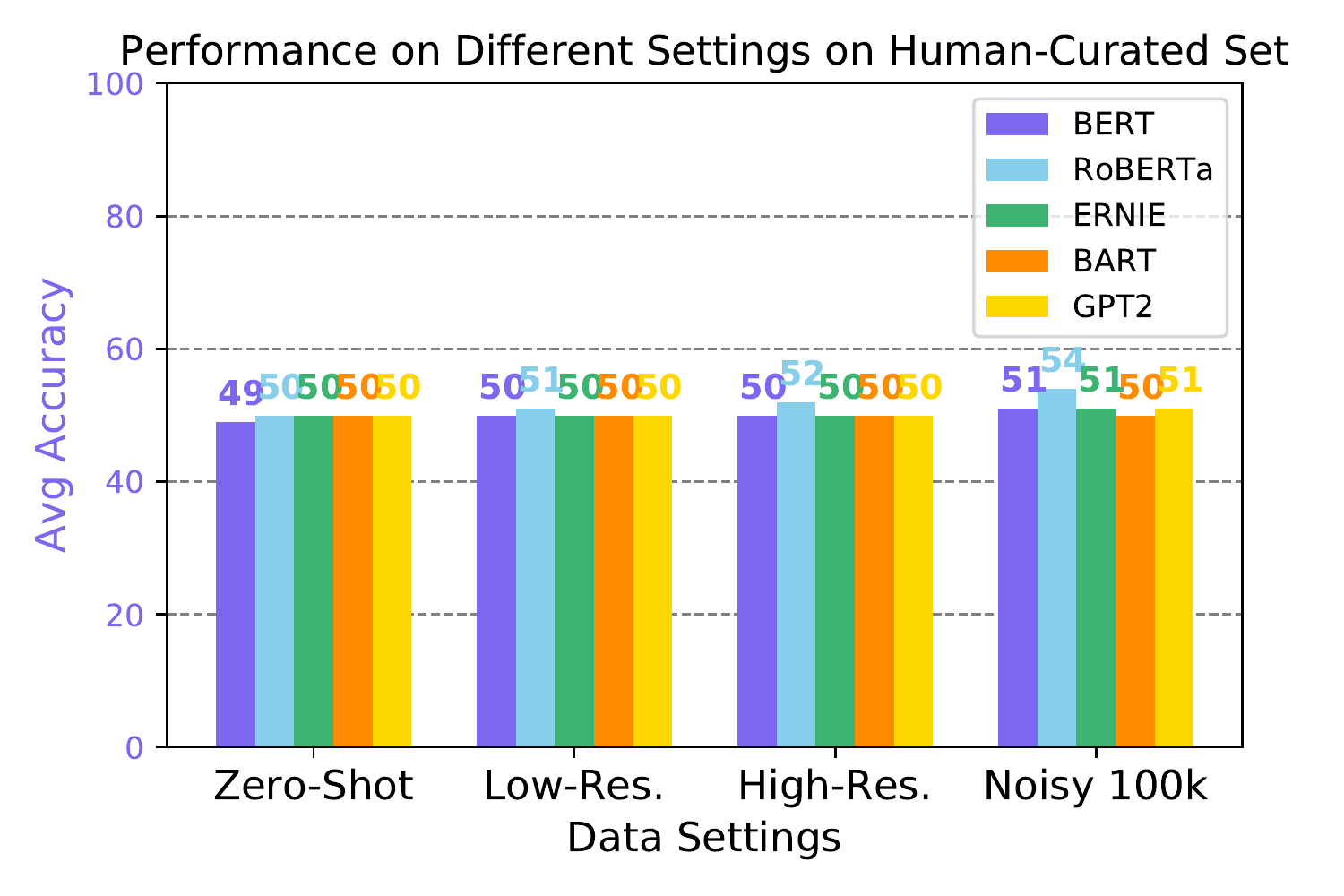}
        \caption{Performance on Human-Curated Set}
        \label{fig:avg_perf_hard}
    \end{subfigure}
    \begin{subfigure}[t]{0.32\linewidth}
    \centering
        \includegraphics[width=\linewidth]{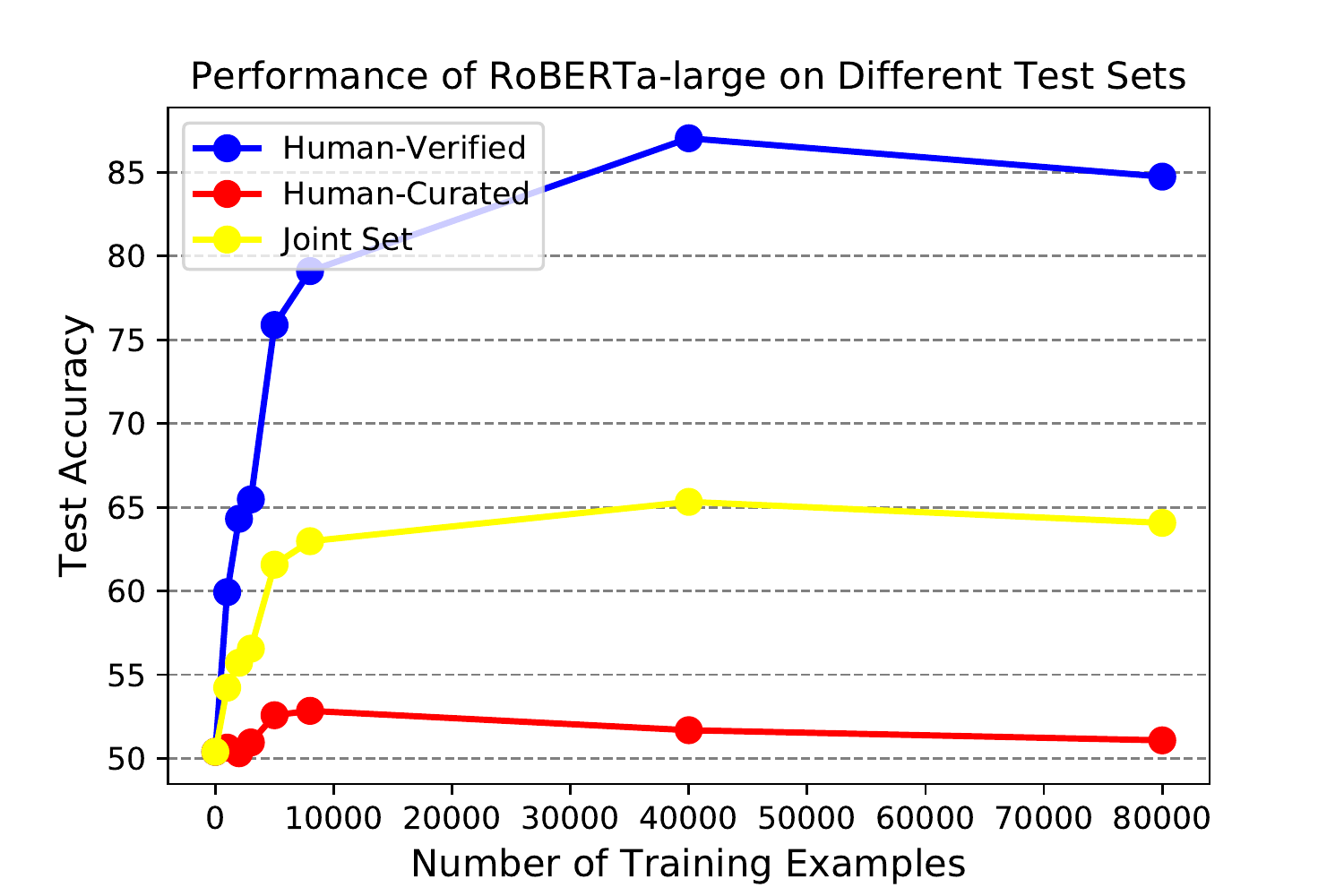}
        \caption{Fine-tuning curve for RoBERTa-large}
        \label{fig:perf_curve}
    \end{subfigure}
    \caption{Performance of different transformer-based models on different settings of our data. BERT, RoBERTa, ERNIE, and BART are evaluated using masked word prediction and GPT2 is evaluated using sentence probability. Zero-shot performance is no better than random guessing. More data helps greatly for human-verified test set (10k) although noisy training hinders the improvement. Increasing data does not help at all for our human-curated set.}
    \label{fig:avg_perf_all_settings}
    \end{figure*}

\para{Human-Verified Set~}
To ensure the quality of crawled data, we conduct human evaluation using Amazon Mechanical Turk (AMT) on 10k of our collected 257k statements covering 1.7k different commonsense axioms. We present a pair of statements by flipping the comparative term in the original statement to its opposite, and ask two annotators to choose the one that follows commonsense. 
If they disagree, we subject the pairs to a second round of turking with three annotators, and use majority voting to validate the statement. When annotators prefer a flipped statement over the original, we replace the statement accordingly. 

\textit{Quality Check.}
The Fleiss-kappa agreement~\cite{fleiss1971measuring} on the two rounds of turking is 0.72 and 0.52, indicating that some statements are difficult for humans to verify. Of 10k statements in the verified set, we sample 10\% (1k with 170 axioms) that 2 annotators agree on in the first or second round to form our \emph{Human-Verified Test Set}.

\para{Human-Curated Set~}
To further challenge models on more flexible forms of text, we ask humans to write perturbed axioms. 
Specifically, given an axiom in FOL, a human annotator is asked to provide input that perturbs the conclusions following \emph{all} 7 types of linguistic perturbations as shown in Table~\ref{tab:LingOperators}, including compositional combinations that are hard to generate using automated approaches, 
Then we apply the asymmetry operators either on the premises or conclusions. Thus we have in total of 24 types of perturbations, including the unperturbed one.
We focus on 80 axioms covering physical, social, and temporal commonsense knowledge and create 1.6k statements. We show examples of all perturbations for one probe in Appendix Table~\ref{tab:perturbations} and sampled 60 probes in Appendix Table~\ref{tab:allprobes}.


\para{Joint Test Set~} Combines the Human-Curated and Human-Verified sets, for a total of 2.6k statements.

\subsection{Evaluation Settings}
\label{subsec: settings}
Using the collected probe data introduced above, we consider four evaluation settings to examine models' capabilities to perform robust inference on our dataset. 

\para{1. Zero-Shot:} In the zero-shot setting, we test models without any exposure to training data. 

\para{2. Low-Resource:} For the low-resource setting, we fine-tune the models on 1k (10\%) of the verified 10k set to determine how a small amount of in-domain text influences PTLM performance.

\para{3. High-Resource:}
We use 90\% of the verified training set (8k for training, 1k for validation). We further increase the number of training instances by introducing 5 different novel entities for each statement, yielding 40k training instances that include 5 repetitions of each probe with different novel entities, providing models more opportunities to learn patterns in the training set.

\para{4. Raw Large-Scale Training:} Finally, to analyze the effects of training on an even larger but noisier set with the similar format. Starting from the raw set of 257k crawled statements, we sample 100k statements from 17k axioms ensuring no overlap with the test set.

\subsection{Baseline Methods}
We evaluate multiple state-of-the-art transformer-based PTLMs covering both masked and generative language models. 
For the masked word prediction task, we consider BERT~\cite{Devlin2019}, RoBERTa~\cite{liu2019roberta}, ERNIE, a knowledge enhanced LM~\cite{zhang2019ernie}, and BART~\cite{lewis2020bart}. 
For sentence probability, we consider GPT-2~\cite{radford2019language}, a unidirectional language model for left-to-right language generation.

\section{Results and Analysis}\label{analysis}
    \begin{figure*}[tb]
    \centering
    \begin{subfigure}[t]{0.32\textwidth}
    \centering
        \includegraphics[width=\textwidth]{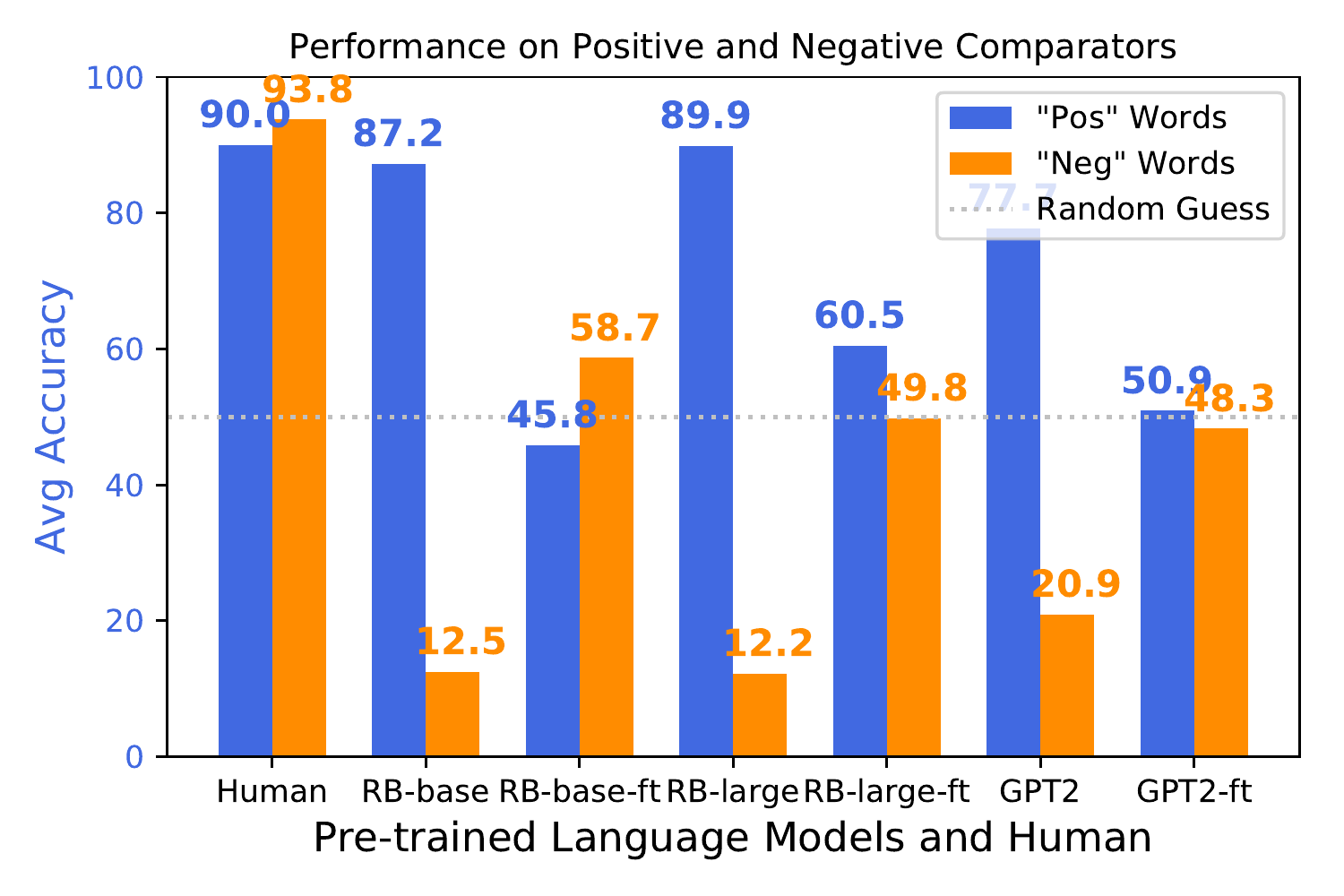}
        \caption{Results on Positivity Bias}
        \label{fig:pos_bias}
    \end{subfigure}%
    \begin{subfigure}[t]{0.32\linewidth}
    \centering
        \includegraphics[width=\linewidth]{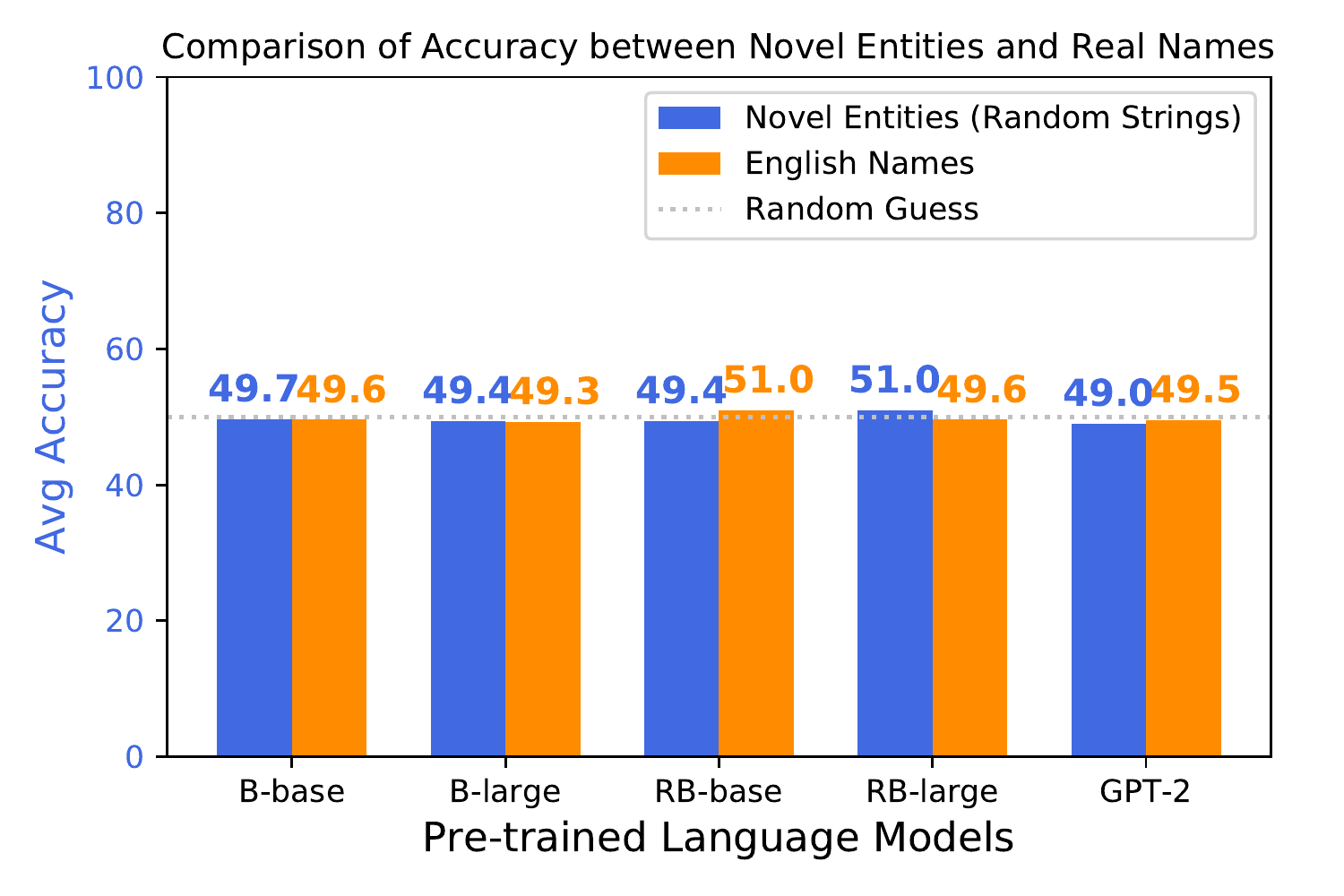}
        \caption{Ablation on Novel Entities}
        \label{fig:ablation}
    \end{subfigure}
    \begin{subfigure}[t]{0.32\linewidth}
    \centering
        \includegraphics[width=\linewidth]{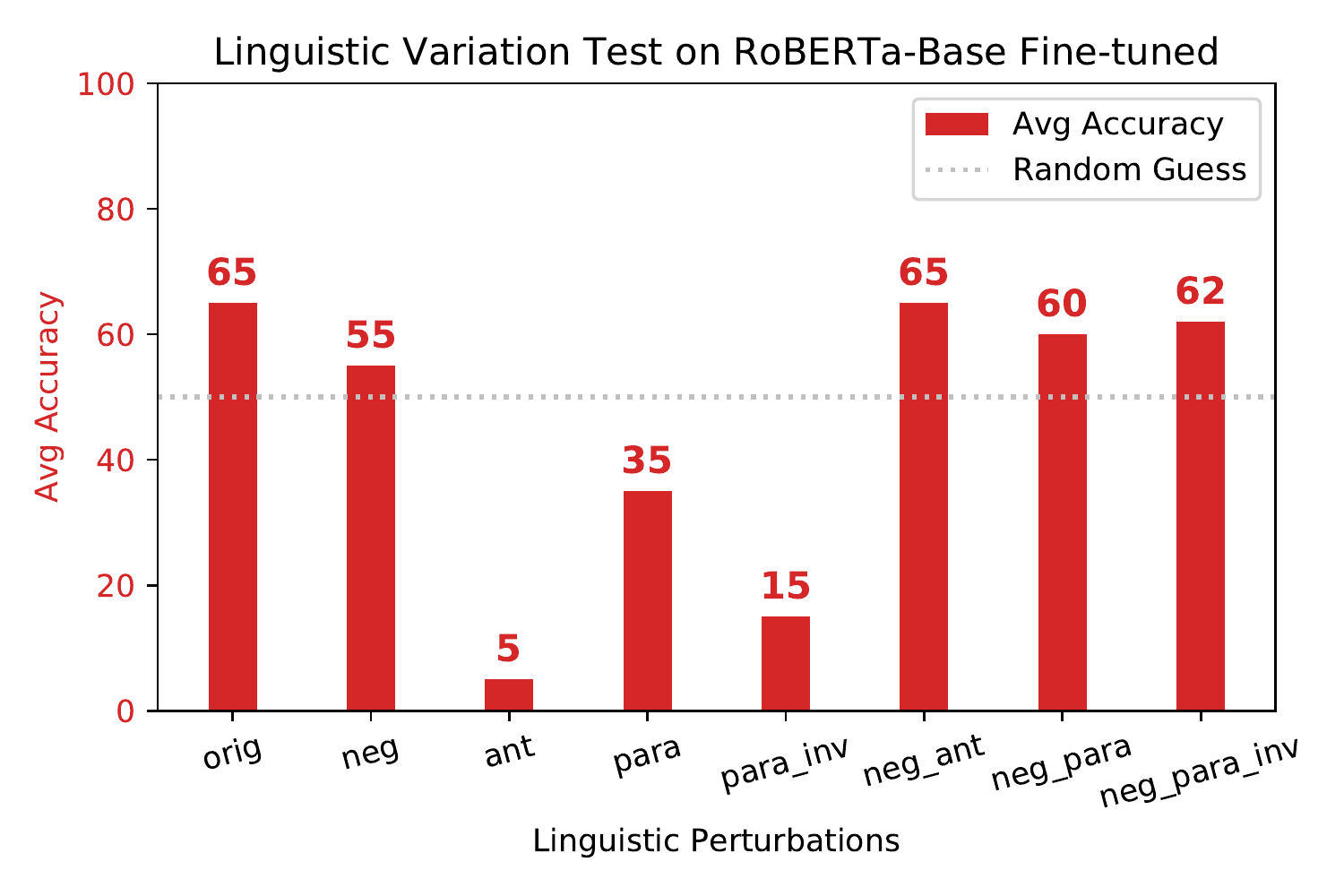}
        \caption{Results per Linguistic Perturbation}
        \label{fig:finetuned_pert}
    \end{subfigure}
    \caption{Results of fine-tuning and the ablation study on novel entities. Shows that (a) models are biased to positive words, requiring fine-tuning to correct (b) poor performance persists after replacing novel entities with real names––indicating the use of random strings is not hindering PTLMs' abilities, (c) fine-tuning mitigates the bias towards positive words, but the inconsistency issue for linguistic variation become obvious.}
    \label{fig:fine-tuning_ablation}
    \end{figure*}
We examine the performance of multiple language models on each evaluation setting on our probe data, including zero-shot and fine-tuning on various splits, and present ablation studies to analyze performance more thoroughly. All of our results are averages of testing on 3 seeds.

\subsection{Zero-Shot Performance}
\label{subsec:exp:no_finetune}




As shown in the first group of bars in Figures~\ref{fig:avg_perf_easy} and~\ref{fig:avg_perf_hard}, the average binary accuracies of all five models (we show the large version) on both MWP and SP tasks are around 0.5, regardless of the test data.
A random baseline that chooses between the two comparative words would have an accuracy of 0.5. This shows that the tested models barely beat a random guessing baseline without training. 

\para{Is Knowledge-Augmented Model Better?}
To see if adding knowledge during training helps, we also test a knowledge-enhanced LM, ERNIE~\cite{zhang2019ernie}. 
However, as we can see in Figures~\ref{fig:avg_perf_easy} and~\ref{fig:avg_perf_hard}, ERNIE also performs on par with random guessing, demonstrating that simply adding more knowledge does not help with the robust inference capability.

\paragraph{Human Performance} 
To benchmark human performance, we sampled 5\% of our joint test set consisting of both human-verified and human-curated data and gathered answers from 20 subjects (annotators) with diverse backgrounds who were not involved in the probe construction process. We consider this as zero-shot testing for humans as they have not seen the training set before.
Humans obtained 91.7\% accuracy, taking a majority vote for each probe, with a substantial inter-annotator agreement of 0.768 Kappa~\cite{cohen1960coefficient}).

\subsection{Fine-tuning Performance}
\label{subsec:exp:finetuned}
To study if poor performance in \subsecref{subsec:exp:no_finetune} is from a lack of exposure to \rica's  probe sets, we conduct experiments to fine-tune baseline language models.
As in~\subsecref{subsec: settings}, we consider training on low-resource data by sampling a subset of the verified set, on high-resource by filling multiple novel entities in the verified set, and the noisy 100k data. 
We fine-tune BERT, RoBERTa, ERNIE, and BART using the same masking approach as MWP evaluation, and fine-tune GPT-2 on the causal language modeling task.
Details for training are in the appendix.

\para{More Data Helps on Human-Verified Set}
Figure~\ref{fig:avg_perf_easy} shows fine-tuning on our probe set helps the model on the human-verified set, especially for RoBERTa and ERNIE, where the high-resource setting almost reaches 90\% accuracy. This demonstrates with enough data, PTLMs are able to reach near-human performance on generated axioms. The low-resource (except for ERNIE) and noisy training settings, however, pose an enduring challenge for most models.

\para{Diversity of Curated Set Stumps All.}
Evaluating models fine-tuned on human-verified data on the human-curated set, where human editors provide greater diversity in probes, tells a different story.  The model accuracy (Figure~\ref{fig:avg_perf_hard}) remains near 50\%, on par with random guessing, for all models in all settings. 
This indicates that exposing these models to numerous linguistically similar sentences does not improve robust inference ability.
Furthermore, we evaluate training data sensitivity for both the human-verified and human-curated set (Figure~\ref{fig:perf_curve}). We vary training set size from 0 to 80k for RoBERTa-large. Our results show that performance on the human-verified set saturates around 80\% accuracy after 10k instances, but human-curated accuracy remains close to 50\% throughout. This casts doubt on the model's generalizability and whether the improved performance may be due to pattern-matching seq2seq generation, not commonsense acquisition.
An inability to improve on reasoning tasks after fine-tuning supports the challenging nature of \rica, which cannot be trivially solved by fine-tuning.
\subsection{Performance Analysis}
\noindent
\textbf{Positivity Bias in PTLMs.}
We find a pattern that when PTLMs are asked to infer a comparative relationship between the property of two entities, the model is heavily biased towards predicting words that evoke positive emotions (positive valence) regardless of what commonsense axiom is embedded in the statement. 
Figure~\ref{fig:pos_bias} shows that the accuracy for ``\textit{positive valence}'' words such as ``\textit{more}'' and ``\textit{easier}'' is much higher than ``\textit{negative valence}'' words such as ``\textit{less}'' and ``\textit{harder}''. 
Fine-tuning on our probes, which have a balanced number of sentences containing positive and negative comparatives, helps mitigate this bias for RoBERTa (base and large) and GPT-2.
We conjecture that this may be due to the frequency difference between positive valence words and negative valence words related to reporting bias in language~\cite{gordon2013reporting}.
~\citet{dodds2015human} shows a universal positivety bias in human languages and to check if our comparators also possess it, we use Google Ngram Viewer~\footnote{\small\url{https://books.google.com/ngrams}} to find frequencies for the masked words, and confirm that the positive valence words are about 5 times more frequent than their negative counterparts.
This correlation supports the claim that PTLMs do not reason as humans do, but are guided by statistical patterns. Our challenge clearly reveals this bias in PTLMs and suggests a potential mitigation using RICA.

\paragraph{Ablation of Novel Entities} 
In order to ensure novel entities used in \rica~did not impact PTLM performance, we conducted an ablation study on 4,800 of our human-curated set (each statement is repeated for 3 times). These probes involved social commonsense, where novel entities took the place of names. We conduct an ablation by choosing common names instead of novel entities, producing probes containing only previously-seen words. As Figure~\ref{fig:ablation} shows, the performance of all models in three settings did not change significantly, strongly suggesting that novel entities are not critical to PTLM performance. We conclude novel entities do not introduce helpful or distracting sub-words.

\paragraph{Impact of Linguistic Perturbations}
Before fine-tuning, a heavy bias for positive valence words interfered with the perturbations analysis, since each perturbation has a balanced number of positive and negative valence words.
After fine-tuning, however, the bias is mitigated and we find significant variations in performance for different perturbation types (Figure~\ref{fig:finetuned_pert}).
This shows that language variation greatly affects a model's capability to make inference on our commonsense probes, while suggesting models do not comprehend the axioms.
Interestingly, the composite perturbation types such as \textsc{Negation Antonym} are not necessarily harder for PTLMs, even though performance on \textsc{Antonym} is the lowest. We speculate that the model is exploiting some pattern in \textsc{Negation Antonym} that is not present for just \textsc{Antonym}.

\section{Related Work}\label{rel_work}

\commentout{
Unfortunately, these two capabilities have largely been overlooked by existing natural language understanding (NLU) benchmarks~\cite{MNLI, levesque2012winograd} and probing studies for transformer-based pre-trained language models ~\cite{vaswani2017attention, Devlin2019, liu2019roberta,clark2020transformers, petroni2019language}.
Most existing natural language inference (NLI) datasets~\cite{MNLI} do not focus on commonsense inferences, while commonsense reasoning benchmarks~\cite{ostermann-etal-2019-commonsense, talmor2019commonsenseqa} do not systematically evaluate robustness against linguistic variations, meaning we cannot preclude the possibility that models are just pattern matching to solve the needed task. 
have a larger scope than just commonsense inference~\cite{MNLI}, thus a strong performance on an NLI dataset does not necessarily imply strong commonsense reasoning abilities. On the other hand commonsense reasoning benchmarks~\cite{zellers2018swagaf,ostermann-etal-2019-commonsense, talmor2019commonsenseqa} do not systematically evaluate robustness against linguistic variations, meaning we cannot preclude the possibility that models are just pattern matching to solve the needed task. 
test inference abilities using known encoded commonsense knowledge~\cite{bisk2019piqa, talmor2019commonsenseqa} and neither of them evaluates robustness on syntactically-diverse probe sets that embed the same axiom.
Neither of them evaluates models' consistency when faced with statements that share the same axiom but are syntactically varied, which is a crucial evidence of understanding the axioms.
As for probing studies, the current success of transformer-based pre-trained language models (PTLMs) such as BERT, RoBERTa, GPT-2, and ALBERT~\cite{Devlin2019, liu2019roberta, radford2019language, lan2019albert} on multiple NLU benchmarks has motivated work that aims to probe different capabilities of PTLMs. 
Besides, 

Previous probing work has shown that PTLMs do posses factual knowledge quite well~\cite{petroni2019language, kwon2019masked} and can emulate deductive reasoning given explicit rules~\cite{clark2020transformers}, but does not probe specifically for the ability to reason robustly with commonsense.}

\para{Commonsense Reasoning} has a long history in AI, with
classical work primarily focusing on executing symbolic rules as hand-crafted programs for machines to learn~\cite{Mccarthy1960ProgramsWC}.
The majority of recent commonsense reasoning benchmarks~\cite{zellers2018swagaf, talmor2019commonsenseqa,bisk2019piqa,sap2019social, lin-etal-2021-riddlesense, lin-etal-2021-xcsr, lin-etal-2021-differentiable, sakaguchi2020winogrande} test a model's ability to choose the correct option given a context and a question; PTLMs have reached high performance on these benchmarks after fine-tuning. 
We differ from these benchmarks by focusing on robustness to linguistic variation via our linguistically-varied commonsense statements. \rica~also challenges PTLMs on two evaluation tasks to better probe the PTLMs' representations.


\para{Reasoning-focused Inference}
There have been many benchmarks that focus on reasoning abilities in multiple tasks such as reading comprehension~\cite{huang2019cosmos, yu2020reclor}, dialogue systems~\cite{cui2020mutual}, and NLI~\cite{MNLI}, that involve inferences on language.
Recent work also aims to probe models in these tasks to see if reasoning is actually achieved~\cite{richardson2020does,richardson2020probing}.

\para{Robustness to Linguistic Variations}
Previous work has also examined model robustness against paraphrases by producing linguistically-varied sentences for different tasks such as NLI~\cite{liu2020empirical,li2020linguistically}, question answering~\cite{weller2020learning}, and sentiment analysis~\cite{ribeiro2018semantically}, just to name a few. Our work distinguishes from them as we look into robustness in regard to commonsense reasoning and develops a systematic procedure to generate paraphrases using first-order logic.

\para{Probing PTLMs} 
Prior works in analyzing the (commonsense) reasoning ability of  PTLMs have primarily focused on creating probing tasks by generating ad-hoc masked sentences either from knowledge bases~\cite{petroni2019language,Feldman2019CommonsenseKM, lin2020numersense} or existing datasets~\cite{zhou2019evaluating, talmor2019olmpics,kwon2019masked,zhou2021probing}.
This first line of works aim to test if PTLMs can work as knowledge bases, i.e. can they retrieve factual knowledge; our work focuses on implicit commonsense relations, not facts.
We differ from the second line of work by proposing a systematic procedure to generate probes and evaluate for robustness.
\citet{clark2020transformers} shows that PTLMs can emulate deductive reasoning given explicit rules, but we focus on unstated commonsense relations.



\section{Conclusion}\label{conclusion}
We design \rica~as an AI challenge to test robust inference capabilities on linguistically-varied probes covering different commonsense axioms.
\rica~is built on a systematic process to construct probes using FOL formulae, perturbation operators, and novel entities.
Following this approach, we generate and verify more than 10k statements from 1.7k axioms and test multiple PTLMs in various settings.
We find that PTLMs perform on par with random guessing on zero-shot setting, have strong positivity bias, and are not robust under linguistic perturbations. 
\section*{Acknowledgments}
We thank anonymous reviewers for providing insightful feedback and members from INK and JAUNTS lab. This research is supported in part by the DARPA MCS program under Contract No. N660011924033, the Defense Advanced Research Projects Agency with award W911NF-19-20271, NSF IIS 2048211, and NSF SMA 182926.

\section*{Ethical Considerations}
Our work aims to pose a new challenge to improve effective human-AI communications by collecting new data in English, which benefits English speakers more. We have conducted human evaluation using Amazon Mechanical Turks. We pay turkers around \$11 per hour, above the national minimum wage and engage in constructive discussions if they have concerns about the process. We also give each annotation instance enough time so that we do not pressure annotators.

Our data construction process makes uses of available public resources: Wikipedia, ConceptNet~\cite{liu2004conceptnet}, and ATOMIC~\cite{sap2019atomic}, which could contain societal biases as shown by~\citet{mehrabi2021lawyers}. Although our probes do not involve specific demographics, we admit the possibility that biases in knowledge resources are included in our data. We have provided detailed descriptions about our data construction process to minimize potential confusions.


\bibliography{emnlp2021}

\begin{thebibliography}{51}
\expandafter\ifx\csname natexlab\endcsname\relax\def\natexlab#1{#1}\fi

\bibitem[{Alshawi and van Eijck(1989)}]{alshawi1989logical}
Hiyan Alshawi and Jan van Eijck. 1989.
\newblock \href {https://doi.org/10.3115/981623.981627} {Logical forms in the
  core language engine}.
\newblock In \emph{27th Annual Meeting of the Association for Computational
  Linguistics}, pages 25--32, Vancouver, British Columbia, Canada. Association
  for Computational Linguistics.

\bibitem[{Bisk et~al.(2020)Bisk, Zellers, LeBras, Gao, and Choi}]{bisk2019piqa}
Yonatan Bisk, Rowan Zellers, Ronan LeBras, Jianfeng Gao, and Yejin Choi. 2020.
\newblock \href {https://aaai.org/ojs/index.php/AAAI/article/view/6239}
  {{PIQA:} reasoning about physical commonsense in natural language}.
\newblock In \emph{The Thirty-Fourth {AAAI} Conference on Artificial
  Intelligence, {AAAI} 2020, The Thirty-Second Innovative Applications of
  Artificial Intelligence Conference, {IAAI} 2020, The Tenth {AAAI} Symposium
  on Educational Advances in Artificial Intelligence, {EAAI} 2020, New York,
  NY, USA, February 7-12, 2020}, pages 7432--7439. {AAAI} Press.

\bibitem[{Bosselut et~al.(2019)Bosselut, Rashkin, Sap, Malaviya, Celikyilmaz,
  and Choi}]{bosselut2019comet}
Antoine Bosselut, Hannah Rashkin, Maarten Sap, Chaitanya Malaviya, Asli
  Celikyilmaz, and Yejin Choi. 2019.
\newblock \href {https://doi.org/10.18653/v1/P19-1470} {{COMET}: Commonsense
  transformers for automatic knowledge graph construction}.
\newblock In \emph{Proceedings of the 57th Annual Meeting of the Association
  for Computational Linguistics}, pages 4762--4779, Florence, Italy.
  Association for Computational Linguistics.

\bibitem[{Carey and Bartlett(1978)}]{carey1978acquiring}
Susan Carey and Elsa Bartlett. 1978.
\newblock Acquiring a single new word.

\bibitem[{Clark and Brennan(1991)}]{clark1991grounding}
Herbert~H Clark and Susan~E Brennan. 1991.
\newblock Grounding in communication.

\bibitem[{Clark et~al.(2020)Clark, Tafjord, and
  Richardson}]{clark2020transformers}
Peter Clark, Oyvind Tafjord, and Kyle Richardson. 2020.
\newblock \href {https://doi.org/10.24963/ijcai.2020/537} {Transformers as soft
  reasoners over language}.
\newblock In \emph{Proceedings of the Twenty-Ninth International Joint
  Conference on Artificial Intelligence, {IJCAI} 2020}, pages 3882--3890.
  ijcai.org.

\bibitem[{Cohen(1960)}]{cohen1960coefficient}
Jacob Cohen. 1960.
\newblock A coefficient of agreement for nominal scales.
\newblock \emph{Educational and psychological measurement}, 20(1):37--46.

\bibitem[{Cui et~al.(2020)Cui, Wu, Liu, Zhang, and Zhou}]{cui2020mutual}
Leyang Cui, Yu~Wu, Shujie Liu, Yue Zhang, and Ming Zhou. 2020.
\newblock \href {https://doi.org/10.18653/v1/2020.acl-main.130} {{M}u{T}ual: A
  dataset for multi-turn dialogue reasoning}.
\newblock In \emph{Proceedings of the 58th Annual Meeting of the Association
  for Computational Linguistics}, pages 1406--1416, Online. Association for
  Computational Linguistics.

\bibitem[{Davison et~al.(2019)Davison, Feldman, and
  Rush}]{Feldman2019CommonsenseKM}
Joe Davison, Joshua Feldman, and Alexander Rush. 2019.
\newblock \href {https://doi.org/10.18653/v1/D19-1109} {Commonsense knowledge
  mining from pretrained models}.
\newblock In \emph{Proceedings of the 2019 Conference on Empirical Methods in
  Natural Language Processing and the 9th International Joint Conference on
  Natural Language Processing (EMNLP-IJCNLP)}, pages 1173--1178, Hong Kong,
  China. Association for Computational Linguistics.

\bibitem[{Devlin et~al.(2019)Devlin, Chang, Lee, and Toutanova}]{Devlin2019}
Jacob Devlin, Ming-Wei Chang, Kenton Lee, and Kristina Toutanova. 2019.
\newblock \href {https://doi.org/10.18653/v1/N19-1423} {{BERT}: Pre-training of
  deep bidirectional transformers for language understanding}.
\newblock In \emph{Proceedings of the 2019 Conference of the North {A}merican
  Chapter of the Association for Computational Linguistics: Human Language
  Technologies, Volume 1 (Long and Short Papers)}, pages 4171--4186,
  Minneapolis, Minnesota. Association for Computational Linguistics.

\bibitem[{Dodds et~al.(2015)Dodds, Clark, Desu, Frank, Reagan, Williams,
  Mitchell, Harris, Kloumann, Bagrow et~al.}]{dodds2015human}
Peter~Sheridan Dodds, Eric~M Clark, Suma Desu, Morgan~R Frank, Andrew~J Reagan,
  Jake~Ryland Williams, Lewis Mitchell, Kameron~Decker Harris, Isabel~M
  Kloumann, James~P Bagrow, et~al. 2015.
\newblock Human language reveals a universal positivity bias.
\newblock \emph{Proceedings of the National Academy of Sciences},
  112(8):2389--2394.

\bibitem[{Fleiss(1971)}]{fleiss1971measuring}
Joseph~L Fleiss. 1971.
\newblock Measuring nominal scale agreement among many raters.
\newblock \emph{Psychological bulletin}, 76(5):378.

\bibitem[{Gordon and Van~Durme(2013)}]{gordon2013reporting}
Jonathan Gordon and Benjamin Van~Durme. 2013.
\newblock Reporting bias and knowledge acquisition.
\newblock In \emph{Proceedings of the 2013 workshop on Automated knowledge base
  construction}, pages 25--30.

\bibitem[{Huang et~al.(2019)Huang, Le~Bras, Bhagavatula, and
  Choi}]{huang2019cosmos}
Lifu Huang, Ronan Le~Bras, Chandra Bhagavatula, and Yejin Choi. 2019.
\newblock \href {https://doi.org/10.18653/v1/D19-1243} {Cosmos {QA}: Machine
  reading comprehension with contextual commonsense reasoning}.
\newblock In \emph{Proceedings of the 2019 Conference on Empirical Methods in
  Natural Language Processing and the 9th International Joint Conference on
  Natural Language Processing (EMNLP-IJCNLP)}, pages 2391--2401, Hong Kong,
  China. Association for Computational Linguistics.

\bibitem[{Kwon et~al.(2019)Kwon, Kang, Han, and Choi}]{kwon2019masked}
Sunjae Kwon, Cheongwoong Kang, Jiyeon Han, and Jaesik Choi. 2019.
\newblock \href {https://arxiv.org/abs/1911.03024} {Why do masked neural
  language models still need common sense knowledge?}
\newblock \emph{ArXiv preprint}, abs/1911.03024.

\bibitem[{Lewis et~al.(2020{\natexlab{a}})Lewis, Liu, Goyal, Ghazvininejad,
  Mohamed, Levy, Stoyanov, and Zettlemoyer}]{lewis2019bart}
Mike Lewis, Yinhan Liu, Naman Goyal, Marjan Ghazvininejad, Abdelrahman Mohamed,
  Omer Levy, Veselin Stoyanov, and Luke Zettlemoyer. 2020{\natexlab{a}}.
\newblock \href {https://doi.org/10.18653/v1/2020.acl-main.703} {{BART}:
  Denoising sequence-to-sequence pre-training for natural language generation,
  translation, and comprehension}.
\newblock In \emph{Proceedings of the 58th Annual Meeting of the Association
  for Computational Linguistics}, pages 7871--7880, Online. Association for
  Computational Linguistics.

\bibitem[{Lewis et~al.(2020{\natexlab{b}})Lewis, Liu, Goyal, Ghazvininejad,
  Mohamed, Levy, Stoyanov, and Zettlemoyer}]{lewis2020bart}
Mike Lewis, Yinhan Liu, Naman Goyal, Marjan Ghazvininejad, Abdelrahman Mohamed,
  Omer Levy, Veselin Stoyanov, and Luke Zettlemoyer. 2020{\natexlab{b}}.
\newblock \href {https://doi.org/10.18653/v1/2020.acl-main.703} {{BART}:
  Denoising sequence-to-sequence pre-training for natural language generation,
  translation, and comprehension}.
\newblock In \emph{Proceedings of the 58th Annual Meeting of the Association
  for Computational Linguistics}, pages 7871--7880, Online. Association for
  Computational Linguistics.

\bibitem[{Li et~al.(2020)Li, Shengshuo, Liu, Wu, Zhou, and
  Steinert-Threlkeld}]{li2020linguistically}
Chuanrong Li, Lin Shengshuo, Zeyu Liu, Xinyi Wu, Xuhui Zhou, and Shane
  Steinert-Threlkeld. 2020.
\newblock Linguistically-informed transformations (lit): A method for
  automatically generating contrast sets.
\newblock In \emph{Proceedings of the Third BlackboxNLP Workshop on Analyzing
  and Interpreting Neural Networks for NLP}, pages 126--135.

\bibitem[{Lin et~al.(2020)Lin, Lee, Khanna, and Ren}]{lin2020numersense}
Bill~Yuchen Lin, Seyeon Lee, Rahul Khanna, and Xiang Ren. 2020.
\newblock \href {https://doi.org/10.18653/v1/2020.emnlp-main.557} {{B}irds have
  four legs?! {N}umer{S}ense: {P}robing {N}umerical {C}ommonsense {K}nowledge
  of {P}re-{T}rained {L}anguage {M}odels}.
\newblock In \emph{Proceedings of the 2020 Conference on Empirical Methods in
  Natural Language Processing (EMNLP)}, pages 6862--6868, Online. Association
  for Computational Linguistics.

\bibitem[{Lin et~al.(2021{\natexlab{a}})Lin, Lee, Qiao, and
  Ren}]{lin-etal-2021-xcsr}
Bill~Yuchen Lin, Seyeon Lee, Xiaoyang Qiao, and Xiang Ren. 2021{\natexlab{a}}.
\newblock \href {https://doi.org/10.18653/v1/2021.acl-long.102} {Common sense
  beyond {E}nglish: Evaluating and improving multilingual language models for
  commonsense reasoning}.
\newblock In \emph{Proceedings of the 59th Annual Meeting of the Association
  for Computational Linguistics and the 11th International Joint Conference on
  Natural Language Processing (Volume 1: Long Papers)}, pages 1274--1287,
  Online. Association for Computational Linguistics.

\bibitem[{Lin et~al.(2021{\natexlab{b}})Lin, Sun, Dhingra, Zaheer, Ren, and
  Cohen}]{lin-etal-2021-differentiable}
Bill~Yuchen Lin, Haitian Sun, Bhuwan Dhingra, Manzil Zaheer, Xiang Ren, and
  William Cohen. 2021{\natexlab{b}}.
\newblock \href {https://doi.org/10.18653/v1/2021.naacl-main.366}
  {Differentiable open-ended commonsense reasoning}.
\newblock In \emph{Proceedings of the 2021 Conference of the North American
  Chapter of the Association for Computational Linguistics: Human Language
  Technologies}, pages 4611--4625, Online. Association for Computational
  Linguistics.

\bibitem[{Lin et~al.(2021{\natexlab{c}})Lin, Wu, Yang, Lee, and
  Ren}]{lin-etal-2021-riddlesense}
Bill~Yuchen Lin, Ziyi Wu, Yichi Yang, Dong-Ho Lee, and Xiang Ren.
  2021{\natexlab{c}}.
\newblock \href {https://doi.org/10.18653/v1/2021.findings-acl.131}
  {{R}iddle{S}ense: Reasoning about riddle questions featuring linguistic
  creativity and commonsense knowledge}.
\newblock In \emph{Findings of the Association for Computational Linguistics:
  ACL-IJCNLP 2021}, pages 1504--1515, Online. Association for Computational
  Linguistics.

\bibitem[{Liu and Singh(2004)}]{liu2004conceptnet}
Hugo Liu and Push Singh. 2004.
\newblock Conceptnet—a practical commonsense reasoning tool-kit.
\newblock \emph{BT technology journal}, 22(4):211--226.

\bibitem[{Liu et~al.(2020)Liu, Xin, Ding, Chang, and Sui}]{liu2020empirical}
Tianyu Liu, Zheng Xin, Xiaoan Ding, Baobao Chang, and Zhifang Sui. 2020.
\newblock An empirical study on model-agnostic debiasing strategies for robust
  natural language inference.
\newblock In \emph{Proceedings of the 24th Conference on Computational Natural
  Language Learning}, pages 596--608.

\bibitem[{Liu et~al.(2019)Liu, Ott, Goyal, Du, Joshi, Chen, Levy, Lewis,
  Zettlemoyer, and Stoyanov}]{liu2019roberta}
Yinhan Liu, Myle Ott, Naman Goyal, Jingfei Du, Mandar Joshi, Danqi Chen, Omer
  Levy, Mike Lewis, Luke Zettlemoyer, and Veselin Stoyanov. 2019.
\newblock \href {https://arxiv.org/abs/1907.11692} {Roberta: A robustly
  optimized bert pretraining approach}.
\newblock \emph{ArXiv preprint}, abs/1907.11692.

\bibitem[{Mccarthy(1960)}]{Mccarthy1960ProgramsWC}
John~W. Mccarthy. 1960.
\newblock Programs with common sense.

\bibitem[{Mehrabi et~al.(2021)Mehrabi, Zhou, Morstatter, Pujara, Ren, and
  Galstyan}]{mehrabi2021lawyers}
Ninareh Mehrabi, Pei Zhou, Fred Morstatter, Jay Pujara, Xiang Ren, and Aram
  Galstyan. 2021.
\newblock \href {https://arxiv.org/abs/2103.11320} {Lawyers are dishonest?
  quantifying representational harms in commonsense knowledge resources}.
\newblock \emph{ArXiv preprint}, abs/2103.11320.

\bibitem[{Ostermann et~al.(2019)Ostermann, Zhang, Roth, and
  Clark}]{ostermann-etal-2019-commonsense}
Simon Ostermann, Sheng Zhang, Michael Roth, and Peter Clark. 2019.
\newblock \href {https://doi.org/10.18653/v1/D19-6007} {Commonsense inference
  in natural language processing ({COIN}) - shared task report}.
\newblock In \emph{Proceedings of the First Workshop on Commonsense Inference
  in Natural Language Processing}, pages 66--74, Hong Kong, China. Association
  for Computational Linguistics.

\bibitem[{Petroni et~al.(2019)Petroni, Rockt{\"a}schel, Riedel, Lewis, Bakhtin,
  Wu, and Miller}]{petroni2019language}
Fabio Petroni, Tim Rockt{\"a}schel, Sebastian Riedel, Patrick Lewis, Anton
  Bakhtin, Yuxiang Wu, and Alexander Miller. 2019.
\newblock \href {https://doi.org/10.18653/v1/D19-1250} {Language models as
  knowledge bases?}
\newblock In \emph{Proceedings of the 2019 Conference on Empirical Methods in
  Natural Language Processing and the 9th International Joint Conference on
  Natural Language Processing (EMNLP-IJCNLP)}, pages 2463--2473, Hong Kong,
  China. Association for Computational Linguistics.

\bibitem[{Radford et~al.()Radford, Wu, Child, Luan, Amodei, and
  Sutskever}]{radford2019language}
Alec Radford, Jeffrey Wu, Rewon Child, David Luan, Dario Amodei, and Ilya
  Sutskever.
\newblock Language models are unsupervised multitask learners.
\newblock \emph{OpenAI Blog 1.8 (2019): 9.}

\bibitem[{Ribeiro et~al.(2018)Ribeiro, Singh, and
  Guestrin}]{ribeiro2018semantically}
Marco~Tulio Ribeiro, Sameer Singh, and Carlos Guestrin. 2018.
\newblock Semantically equivalent adversarial rules for debugging nlp models.
\newblock In \emph{Proceedings of the 56th Annual Meeting of the Association
  for Computational Linguistics (Volume 1: Long Papers)}, pages 856--865.

\bibitem[{Richardson et~al.(2020)Richardson, Hu, Moss, and
  Sabharwal}]{richardson2020probing}
Kyle Richardson, Hai Hu, Lawrence~S Moss, and Ashish Sabharwal. 2020.
\newblock Probing natural language inference models through semantic fragments.
\newblock In \emph{AAAI}, pages 8713--8721.

\bibitem[{Richardson and Sabharwal(2020)}]{richardson2020does}
Kyle Richardson and Ashish Sabharwal. 2020.
\newblock \href {https://doi.org/10.1162/tacl_a_00331} {What does my {QA} model
  know? devising controlled probes using expert knowledge}.
\newblock \emph{Transactions of the Association for Computational Linguistics},
  8:572--588.

\bibitem[{Sakaguchi et~al.(2020)Sakaguchi, Le~Bras, Bhagavatula, and
  Choi}]{sakaguchi2020winogrande}
Keisuke Sakaguchi, Ronan Le~Bras, Chandra Bhagavatula, and Yejin Choi. 2020.
\newblock Winogrande: An adversarial winograd schema challenge at scale.
\newblock In \emph{Proceedings of the AAAI Conference on Artificial
  Intelligence}, volume~34, pages 8732--8740.

\bibitem[{Sap et~al.(2019{\natexlab{a}})Sap, Bras, Allaway, Bhagavatula,
  Lourie, Rashkin, Roof, Smith, and Choi}]{sap2019atomic}
Maarten Sap, Ronan~Le Bras, Emily Allaway, Chandra Bhagavatula, Nicholas
  Lourie, Hannah Rashkin, Brendan Roof, Noah~A. Smith, and Yejin Choi.
  2019{\natexlab{a}}.
\newblock \href {https://doi.org/10.1609/aaai.v33i01.33013027} {{ATOMIC:} an
  atlas of machine commonsense for if-then reasoning}.
\newblock In \emph{The Thirty-Third {AAAI} Conference on Artificial
  Intelligence, {AAAI} 2019, The Thirty-First Innovative Applications of
  Artificial Intelligence Conference, {IAAI} 2019, The Ninth {AAAI} Symposium
  on Educational Advances in Artificial Intelligence, {EAAI} 2019, Honolulu,
  Hawaii, USA, January 27 - February 1, 2019}, pages 3027--3035. {AAAI} Press.

\bibitem[{Sap et~al.(2019{\natexlab{b}})Sap, Rashkin, Chen, Le~Bras, and
  Choi}]{sap2019social}
Maarten Sap, Hannah Rashkin, Derek Chen, Ronan Le~Bras, and Yejin Choi.
  2019{\natexlab{b}}.
\newblock \href {https://doi.org/10.18653/v1/D19-1454} {Social {IQ}a:
  Commonsense reasoning about social interactions}.
\newblock In \emph{Proceedings of the 2019 Conference on Empirical Methods in
  Natural Language Processing and the 9th International Joint Conference on
  Natural Language Processing (EMNLP-IJCNLP)}, pages 4463--4473, Hong Kong,
  China. Association for Computational Linguistics.

\bibitem[{Schank and Abelson(1977)}]{schank1977scripts}
Roger~C Schank and Robert~P Abelson. 1977.
\newblock Scripts, plans, goals and understanding: An inquiry into human
  knowledge structures.

\bibitem[{Smilkov et~al.(2017)Smilkov, Thorat, Kim, Vi{\'e}gas, and
  Wattenberg}]{smilkov2017smoothgrad}
Daniel Smilkov, Nikhil Thorat, Been Kim, Fernanda Vi{\'e}gas, and Martin
  Wattenberg. 2017.
\newblock Smoothgrad: removing noise by adding noise.
\newblock \emph{ICML Workshop Workshop on Visualization for Deep Learning.}

\bibitem[{Talmor et~al.(2020)Talmor, Elazar, Goldberg, and
  Berant}]{talmor2019olmpics}
Alon Talmor, Yanai Elazar, Yoav Goldberg, and Jonathan Berant. 2020.
\newblock \href {https://doi.org/10.1162/tacl_a_00342} {o{LM}pics-on what
  language model pre-training captures}.
\newblock \emph{Transactions of the Association for Computational Linguistics},
  8:743--758.

\bibitem[{Talmor et~al.(2019)Talmor, Herzig, Lourie, and
  Berant}]{talmor2019commonsenseqa}
Alon Talmor, Jonathan Herzig, Nicholas Lourie, and Jonathan Berant. 2019.
\newblock \href {https://doi.org/10.18653/v1/N19-1421} {{C}ommonsense{QA}: A
  question answering challenge targeting commonsense knowledge}.
\newblock In \emph{Proceedings of the 2019 Conference of the North {A}merican
  Chapter of the Association for Computational Linguistics: Human Language
  Technologies, Volume 1 (Long and Short Papers)}, pages 4149--4158,
  Minneapolis, Minnesota. Association for Computational Linguistics.

\bibitem[{Vaswani et~al.(2017)Vaswani, Shazeer, Parmar, Uszkoreit, Jones,
  Gomez, Kaiser, and Polosukhin}]{vaswani2017attention}
Ashish Vaswani, Noam Shazeer, Niki Parmar, Jakob Uszkoreit, Llion Jones,
  Aidan~N. Gomez, Lukasz Kaiser, and Illia Polosukhin. 2017.
\newblock \href
  {https://proceedings.neurips.cc/paper/2017/hash/3f5ee243547dee91fbd053c1c4a845aa-Abstract.html}
  {Attention is all you need}.
\newblock In \emph{Advances in Neural Information Processing Systems 30: Annual
  Conference on Neural Information Processing Systems 2017, December 4-9, 2017,
  Long Beach, CA, {USA}}, pages 5998--6008.

\bibitem[{Wallace et~al.(2019)Wallace, Tuyls, Wang, Subramanian, Gardner, and
  Singh}]{wallace2019allennlp}
Eric Wallace, Jens Tuyls, Junlin Wang, Sanjay Subramanian, Matt Gardner, and
  Sameer Singh. 2019.
\newblock \href {https://doi.org/10.18653/v1/D19-3002} {{A}llen{NLP} interpret:
  A framework for explaining predictions of {NLP} models}.
\newblock In \emph{Proceedings of the 2019 Conference on Empirical Methods in
  Natural Language Processing and the 9th International Joint Conference on
  Natural Language Processing (EMNLP-IJCNLP): System Demonstrations}, pages
  7--12, Hong Kong, China. Association for Computational Linguistics.

\bibitem[{Weller et~al.(2020)Weller, Lourie, Gardner, and
  Peters}]{weller2020learning}
Orion Weller, Nicholas Lourie, Matt Gardner, and Matthew Peters. 2020.
\newblock Learning from task descriptions.
\newblock In \emph{Proceedings of the 2020 Conference on Empirical Methods in
  Natural Language Processing (EMNLP)}, pages 1361--1375.

\bibitem[{Williams et~al.(2018)Williams, Nangia, and Bowman}]{MNLI}
Adina Williams, Nikita Nangia, and Samuel Bowman. 2018.
\newblock \href {https://doi.org/10.18653/v1/N18-1101} {A broad-coverage
  challenge corpus for sentence understanding through inference}.
\newblock In \emph{Proceedings of the 2018 Conference of the North {A}merican
  Chapter of the Association for Computational Linguistics: Human Language
  Technologies, Volume 1 (Long Papers)}, pages 1112--1122, New Orleans,
  Louisiana. Association for Computational Linguistics.

\bibitem[{Yu et~al.(2020)Yu, Jiang, Dong, and Feng}]{yu2020reclor}
Weihao Yu, Zihang Jiang, Yanfei Dong, and Jiashi Feng. 2020.
\newblock \href {https://openreview.net/forum?id=HJgJtT4tvB} {Reclor: {A}
  reading comprehension dataset requiring logical reasoning}.
\newblock In \emph{8th International Conference on Learning Representations,
  {ICLR} 2020, Addis Ababa, Ethiopia, April 26-30, 2020}. OpenReview.net.

\bibitem[{Zellers et~al.(2018)Zellers, Bisk, Schwartz, and
  Choi}]{zellers2018swagaf}
Rowan Zellers, Yonatan Bisk, Roy Schwartz, and Yejin Choi. 2018.
\newblock \href {https://doi.org/10.18653/v1/D18-1009} {{SWAG}: A large-scale
  adversarial dataset for grounded commonsense inference}.
\newblock In \emph{Proceedings of the 2018 Conference on Empirical Methods in
  Natural Language Processing}, pages 93--104, Brussels, Belgium. Association
  for Computational Linguistics.

\bibitem[{Zhang et~al.(2017)Zhang, Rudinger, Duh, and
  Van~Durme}]{zhang2017ordinal}
Sheng Zhang, Rachel Rudinger, Kevin Duh, and Benjamin Van~Durme. 2017.
\newblock \href {https://doi.org/10.1162/tacl_a_00068} {Ordinal common-sense
  inference}.
\newblock \emph{Transactions of the Association for Computational Linguistics},
  5:379--395.

\bibitem[{Zhang et~al.(2019)Zhang, Han, Liu, Jiang, Sun, and
  Liu}]{zhang2019ernie}
Zhengyan Zhang, Xu~Han, Zhiyuan Liu, Xin Jiang, Maosong Sun, and Qun Liu. 2019.
\newblock \href {https://doi.org/10.18653/v1/P19-1139} {{ERNIE}: Enhanced
  language representation with informative entities}.
\newblock In \emph{Proceedings of the 57th Annual Meeting of the Association
  for Computational Linguistics}, pages 1441--1451, Florence, Italy.
  Association for Computational Linguistics.

\bibitem[{Zhou et~al.(2021{\natexlab{a}})Zhou, Gopalakrishnan, Hedayatnia, Kim,
  Pujara, Ren, Liu, and Hakkani-Tur}]{zhou-etal-2021-commonsense}
Pei Zhou, Karthik Gopalakrishnan, Behnam Hedayatnia, Seokhwan Kim, Jay Pujara,
  Xiang Ren, Yang Liu, and Dilek Hakkani-Tur. 2021{\natexlab{a}}.
\newblock \href {https://aclanthology.org/2021.sigdial-1.13}
  {Commonsense-focused dialogues for response generation: An empirical study}.
\newblock In \emph{Proceedings of the 22nd Annual Meeting of the Special
  Interest Group on Discourse and Dialogue}, pages 121--132, Singapore and
  Online. Association for Computational Linguistics.

\bibitem[{Zhou et~al.(2021{\natexlab{b}})Zhou, Jandaghi, Lin, Cho, Pujara, and
  Ren}]{zhou2021probing}
Pei Zhou, Pegah Jandaghi, Bill~Yuchen Lin, Justin Cho, Jay Pujara, and Xiang
  Ren. 2021{\natexlab{b}}.
\newblock Probing causal common sense in dialogue response generation.
\newblock \emph{In Findings of EMNLP}.

\bibitem[{Zhou et~al.(2020)Zhou, Zhang, Cui, and Huang}]{zhou2019evaluating}
Xuhui Zhou, Yue Zhang, Leyang Cui, and Dandan Huang. 2020.
\newblock Evaluating commonsense in pre-trained language models.
\newblock \emph{AAAI}.

\end{thebibliography}
\bibliographystyle{acl_natbib}
\clearpage
\appendix
\label{appendix}


\section{Probing Data Details}
\label{data_details}
\subsection{Raw Set Collection}
We define 1-4 combinations of type constraints on the predicates for each LT and designe crawling strategies accordingly using resources: ConceptNet, ATOMIC, and Wikipedia. Descriptions for each of the 11 strategies are included in Table~\ref{tab:strategies}. All data and code for crawling strategies is included in the supplementary materials.

\subsection{Turking Details for Human-Verified Set}
We present a pair of statements by flipping the comparative term in the original statement to its opposite, and ask two annotators to choose the one that follows commonsense. The AMT page for turkers to annotate is shown in Figure~\ref{fig:turking}.
If they disagree, we then take the pairs and do a second round of turking by asking three annotators and use majority voting to decide what is the right sentence in the pair. We replace the original statement with the opposite one if there are more annotators think that the other one in the pair follows more commonsense. In total, around 2500 pairs are sent to the second round and 300 pairs are flipped to the opposite according to annotators. The estimated time for completing each instance is around 20 seconds and we pay each instance \$0.06, which translates to around \$11 per hour.

\label{turking_details}
\subsection{Human-Curated Set Details}
We show all perturbations for one probe in Table~\ref{tab:perturbations} and 60 of our human-curated set's unperturbed statement in Table~\ref{tab:allprobes} (for temporal refer to supplementary material). Full data is included in the supplementary material.

\begin{table*}[]
\resizebox{\textwidth}{!}{%
\begin{tabular}{|c|l|l|l|}
\hline
\textbf{Logical Template} & \multicolumn{1}{c|}{\textbf{Type Constraint}} & \multicolumn{1}{c|}{\textbf{Crawling Strategy}}                                                                                                                                                                                                            & \multicolumn{1}{c|}{\textbf{Example Axiom (Adjusted for readability)}}                                                                                           \\ \hline
\multirow{4}{*}{1}        & Attribute-Material (126)                                & \begin{tabular}[c]{@{}l@{}}Get a list of materials, and find properties in ConceptNet using HasProperty;\\ then find a second material using NotMadeOf from the previous property.\end{tabular}                                                            & Material(A, glass) and Material(B, wood), so More(clear(A), clear(B))                                                                                            \\ \cline{2-4} 
                          & Attribute-Grade (132)                                   & \begin{tabular}[c]{@{}l@{}}Input an ordered list of numbers and form pairs and comparative relations \\ following the orders\end{tabular}                                                                                                                  & Grade(A, first) and Grade(B, third), so More(young(A), young(B))                                                                                                 \\ \cline{2-4} 
                          & Condition-Location (1k)                                 & \begin{tabular}[c]{@{}l@{}}Get a list of places with descending latitude from Wikipedia and form pairs\\ by the relation (higher latitude is colder than lower latitude), considering \\ both hemispheres.\end{tabular}                                    & \begin{tabular}[c]{@{}l@{}}Location(A,equator) and Location(B, north pole), so \\ More(living in hot condition(A), living in hot condition(B))\end{tabular}      \\ \cline{2-4} 
                          & Attribute-Animal (10k)                                  & \begin{tabular}[c]{@{}l@{}}Get a list of animals from Wikipedia and find properties in ConceptNet \\ using CapableOf and LocateAt\end{tabular}                                                                                                             & \begin{tabular}[c]{@{}l@{}}Animal(A, fish) and Animal(B, horse), so More(locate at the bottom \\ of the sea(A), locate at the bottom of the sea(B))\end{tabular} \\ \hline
\multirow{2}{*}{2}        & Role (1.2k)                                   & \begin{tabular}[c]{@{}l@{}}Input a list of occupations from Wikipedia and find properties in ConceptNet\\ using CapableOf\end{tabular}                                                                                                                     & Priest(A,B), so More(Pray(A), Pray(B))                                                                                                                           \\ \cline{2-4} 
                          & Action (10k)                                  & \begin{tabular}[c]{@{}l@{}}For every event in ATOMIC that involves two people, we find properties \\ by following the Attribute edge in ATOMIC\end{tabular}                                                                                                & Forces upon(A, B), so More(pushy(A), pushy(B))                                                                                                                   \\ \hline
\multirow{2}{*}{3}        & Action (10k)                                   & \begin{tabular}[c]{@{}l@{}}For each event in ATOMIC that involves people, we find properties \\ by following the Attribute edge in ATOMIC, note that we replace PersonX \\ with "themself" and PersonY with "another person" to sound natural\end{tabular} & \begin{tabular}[c]{@{}l@{}}Assesses patient(A) and not Assesses patient(B), \\ so More(analytical(A), analytical(B))\end{tabular}                                \\ \cline{2-4} 
                          & Capability-Physical (100)                                & \begin{tabular}[c]{@{}l@{}}Input a list of adjectives describing objects, we find properties by following\\  UsedFor edge in ConceptNet\end{tabular}                                                                                                       & Tie knot(A) and not Tie knot(B), so More(elastic(A), elastic(B))                                                                                                 \\ \hline
\multirow{2}{*}{4}        & Action (10k)                                  & Similarly to LT3-Event                                                                                                                                                                                                                                     & More(Concentrate(A), Concentrate(B)), so More(Effective(A), Effective(B))                                                                                        \\ \cline{2-4} 
                          & Capability-Physical (100)                                & Simiarly to LT3-Physical                                                                                                                                                                                                                                   & \begin{tabular}[c]{@{}l@{}}More(square(A), square(B)), so\\  Better(divide two space(A), divide two space(B))\end{tabular}                                       \\ \hline
5                         & Attribute-Temporal (100)                                & \begin{tabular}[c]{@{}l@{}}Manually come up with temporal ordered-events, included in Human-Curated\\ Set\end{tabular}                                                                                                                                     & entered the building(A),so before(outside(A))                                                                                                                    \\ \hline
\end{tabular}
}
\caption{Crawling strategies for 11 type-constrained KT crawling for our Raw Set.}
\label{tab:strategies}
\end{table*}

\section{Experimental Details}
\para{Model Detail}
We test our probes on in total 10 models, with the number of parameters and other details in Table~\ref{tab:models}.
For RoBERTa-base, RoBERTa-large, RoBERTa-large-MNLI, and BART-large-MNLI, we use the fairseq implementation~\footnote{\url{https://github.com/pytorch/fairseq/tree/master/examples/roberta}, \url{https://github.com/pytorch/fairseq/tree/master/examples/bart}}.
For BERT-base-uncased, BERT-large-uncased, ALBERT, and GPT-2, we use the huggingface transformers library~\footnote{\url{https://huggingface.co/transformers/model_doc/albert.html}, \url{https://huggingface.co/transformers/model_doc/gpt2.html}}.
For COMET trained on ConceptNet and ATOMIC, we follow their github repo~\footnote{\url{https://github.com/atcbosselut/comet-commonsense}}. We use ERNIE from their original github~\footnote{\url{https://github.com/thunlp/ERNIE}}.

\para{Fine-tuning Details}
We fine-tune BERT-base-uncased, BERT-large-uncased, RoBERTa-base, and RoBERTa-large based on HappyTransformers~\footnote{\url{https://github.com/EricFillion/happy-transformer}} framework, using a consistent learning rate of 1e-5.
We fine-tune GPT-2 based on huggingface transformers library's example code~\footnote{\url{https://github.com/huggingface/transformers/tree/master/examples/language-modeling}}, using their default parameters.
We train them on one NVIDIA Quadro RTX 6000 GPU for 10 epochs and after each epoch we test the fine-tuned model on our validation set, and save the model with the highest validation set performance.
Fine-tuning RoBERTa-base and GPT-2 takes around 30 minutes for each epoch and RoBERTa-large takes around 1 hour.
The best validation performance for RoBERTa-base is the fourth epoch, with perplexity 1.3378140926361084 and evaluation loss': 0.2910370217429267.
For RoBERTa-large, the best is epoch 5, with perplexity 1.3949965238571167 and evaluation loss 0.3328918993473053.
For GPT-2, the best is epoch 3, with perplexity 1.2786548795017285.

\para{Interpretation Details}
We use the AllenInterpret demo~\footnote{\url{https://demo.allennlp.org/masked-lm}}.
To identify important context words, we run the algorithm over the same probe for 5 times, each with different entity names, and select the words that are ranked in the top 5 most important words at least 3 times.
We find that the interpretations are not very consistent as the most important words change when we input the same sentence for multiple times and will also change when different names are used, so we conduct 5 trials with different names for each probe and pick the words that appear in the majority of the trials. 

\begin{table}[t]
\centering
\scalebox{0.65}{
\begin{tabular}{l|l}
\toprule
\textsc{\textbf{Category}}        & \multicolumn{1}{c}{\textsc{\textbf{Example}}}                                                                            \\ \midrule
Physical (30\%) & {\begin{tabular}[c]{@{}l@{}}A is smaller than B, \\ so A is easier to put into a box than B.\end{tabular}} \\ \midrule
Material (30\%) & \begin{tabular}[c]{@{}l@{}}A is made out of glass and B is made out of stone, \\ so A is more transparent than B.\end{tabular}  \\ \midrule
Social (30\%)   & \begin{tabular}[c]{@{}l@{}}A makes the varsity team while B does not, \\ so A is more skilled than B.\end{tabular}              \\ \midrule
Temporal (10\%) & \begin{tabular}[c]{@{}l@{}}A was eating dinner, \\ so A was hungry before eating dinner.\end{tabular}                           \\ \bottomrule
\end{tabular}
}
\caption{Different types of commonsense axioms included in our human-curated probe set}
\label{tab:KnowledgeType}
\end{table}

\section{Additional Studies}
\paragraph{Does explicitly providing commonsense knowledge help?}
Shocked by the severe bias observed in PTLMs, we construct an easier set of probes, where we explicitly state all knowledge needed to make the correct logical inference. We have two settings for this test, one where parroting the now-provided commonsense fact is all that is needed to correctly answer the probe, and the other where a simple negation switch of the commonsense fact is needed to solve the probe:
\begin{itemize}
\item A is made of glass, B is made of stone, \textit{and glass is more transparent than stone}, so A is [MASK] transparent than stone. (parrot)
\item A is made of glass, B is made of stone, \textit{and glass is more transparent than stone}, so A is \textbf{not} [MASK] transparent than stone. (negation switch)
\end{itemize}

\begin{table}[]
\centering
\small
\scalebox{0.8}{
\begin{tabular}{|c|c|}
\hline
Model          & Details                                                                                                              \\ \hline
BERT-base-uncased   & 12-layer, 768-hidden, 12-heads, 125M parameters                                                                      \\ \hline
BERT-large-uncased  & 24-layer, 1024-hidden, 16-heads, 355M parameters                                                                     \\ \hline
RoBERTa-base   & 12-layer, 768-hidden, 12-heads, 125M parameters                                                                      \\ \hline
RoBERTa-large  & 24-layer, 1024-hidden, 16-heads, 355M parameters                                                                     \\ \hline
ALBERT         & \begin{tabular}[c]{@{}c@{}}12 repeating layer, 128 embedding, \\ 4096-hidden, 64-heads, 223M parameters\end{tabular} \\ \hline
GPT-2         & 12-layer, 768-hidden, 12-heads, 117M parameters.                                                                     \\ \hline
COMET-Concept  & GPT-2 config + Traning on ConceptNet                                                                                 \\ \hline
COMET-ATOMIC  & GPT-2 config + Traning on ATOMIC                                                                                     \\ \hline
RoBERTa-L-MNLI & 24-layer, 1024-hidden, 16-heads, 355M parameters                                                                     \\ \hline
BART-L-MNLI    & \begin{tabular}[c]{@{}c@{}}24-layer, 1024-hidden, 16-heads, 406M parameters\\ + a classification head\end{tabular}   \\ \hline
\end{tabular}
}
\caption{Models tested and details.}
\label{tab:models}
\end{table}

We do this so to investigate whether RoBERTa is actually able to use the provided commonsense fact, or is it possibly just pattern matching. 

\begin{figure}
	\centering
	\includegraphics[width=\linewidth]{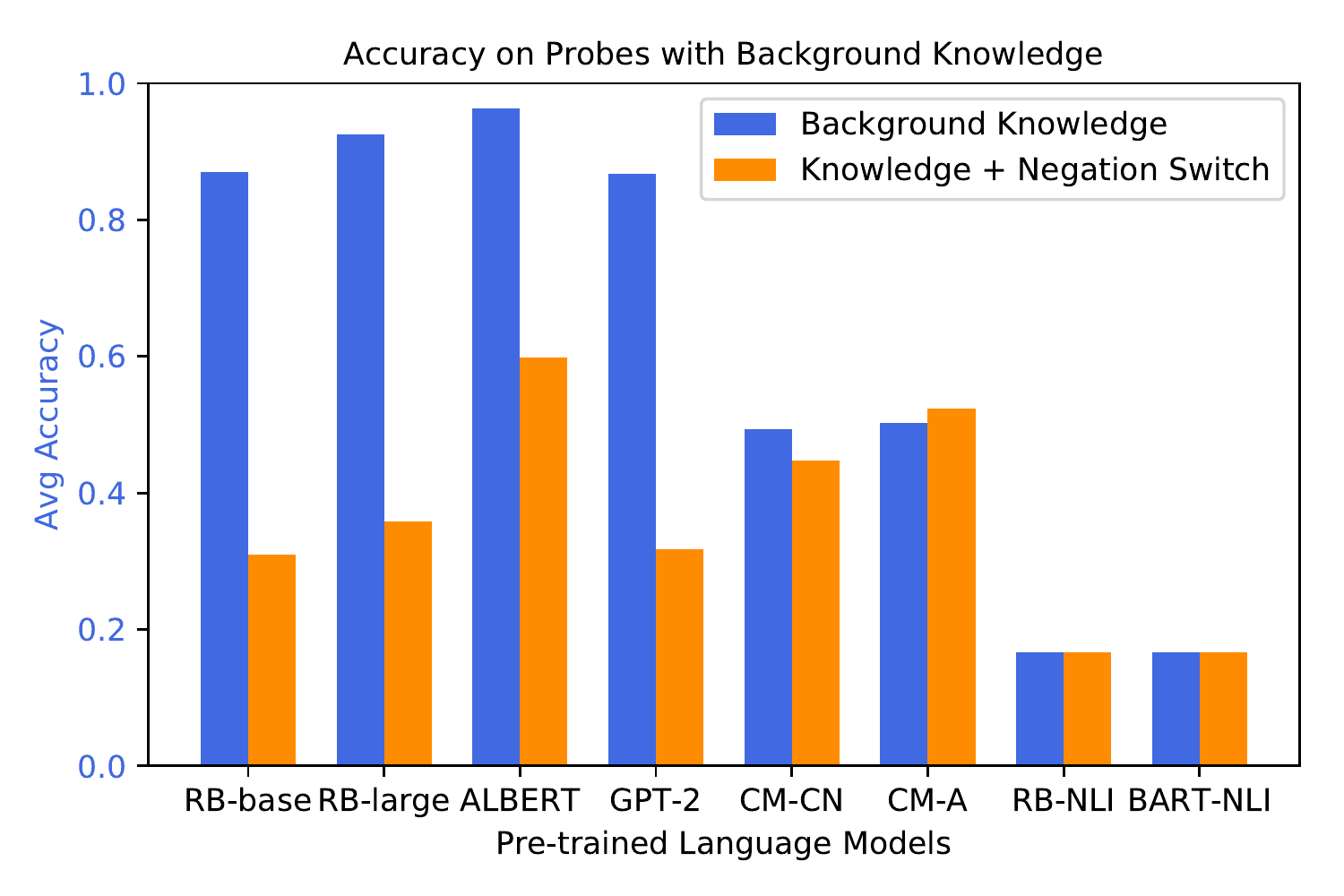}
	\caption{\small{Results of average performance of PTLMs when we provide background knowledge in our probes. For RoBERTa, ALBERT, and GPT-2, we notice a huge increase in accuracy when provided knowledge. However, we find that they are merely parroting what appears in the context since when we apply a negation in the probe, which should change the prediction, they are simply predicting the same as the context shows, resulting in performance drop. For COMET moddels and models tested on the NLI setting, we do not observe the same pattern and it seems that adding knowledge does not help or hurt.}}
	\label{fig:background_knowledge}
\end{figure}

We add this piece of background knowledge to the 60 original (unperturbed) statements along with their corresponding negated statements to form an ``easier'' setting of our task. As shown in Figure~\ref{fig:background_knowledge}, we find two patterns PTLMs exhibit.
For RoBERTa, ALBERT, and GPT-2, there is a stark difference in performance between the two settings. When they are being asked to parrot the commonsense fact, the performances jump up to near perfect scores, however when all they have to do is the equivalent of applying a negation operator on the fact, they fail even worse than when they are not provided the fact. 
These results suggest that in the parrot easier setting, it is likely RoBERTa, ALBERT, and GPT-2 are just parroting the commonsense fact they see in the sentence and not utilizing some sort of reasoning ability, as when asked to perform the simplest of logical operations they fail.
The other pattern we notice is that providing background knowledge does not help or hurt the performances for COMET and models tested on the textual entailment task.
For COMET models, this may be due to the fact that COMET is trained on triplets from knowledge bases: given a head entity and a relation, predict the tail entity, so it is not used to taking auxiliary knowledge into its input.
As for models fine-tuned on MNLI, the performance stays unchanged because they still think most of the sentence pairs of our probes are neutral, failing to grasp the embedded logical inference step.

\para{Case Study on Contextual Clues}
To gain a better understanding on model behaviors, we conduct analysis to identify context words that the model relies on when solving our probes.
We use the SmoothGrad~\cite{smilkov2017smoothgrad} algorithm from AllenNLP Interpret~\cite{wallace2019allennlp} for masked word prediction on our probes with real people's names (the same set as our ablation study) using BERT.
Aggregated across all probe sets, we find that the three words BERT finds most important are: ``\textit{than}'', ``\textit{not}'', and ``\textit{so}'', which make sense as they are indicators for comparison, negation, and causality, respectively.

``\textit{Not}'' and ``\textit{so}'' are also the textual forms of the logical connectives $\neg$ and $\rightarrow$, which we use to construct LTs.

Furthermore, we find that BERT also regards \emph{argument} words (inputs into LTs' predicates via a knowledge table, such as ``\textit{lawyer}'' or ``\textit{knowledge of law}'') important.
The model finds on average 3.4 words as contextual clues and 1.5 out of them are knowledge-specific argument words.
This finding shows that a PTLM is able to recognize words specific to the commonsense axiom tested. 
However, noticing all these clues does not necessarily aid in a PTLM's ability to understand their logical implications, as evidenced by their performances.
In other words, a PTLM, in this case BERT, knows that these words are important when making a decision, but it does not know how to properly answer \rica's questions based on these lexical signals.

\begin{figure}
	\centering
	\includegraphics[width=\linewidth]{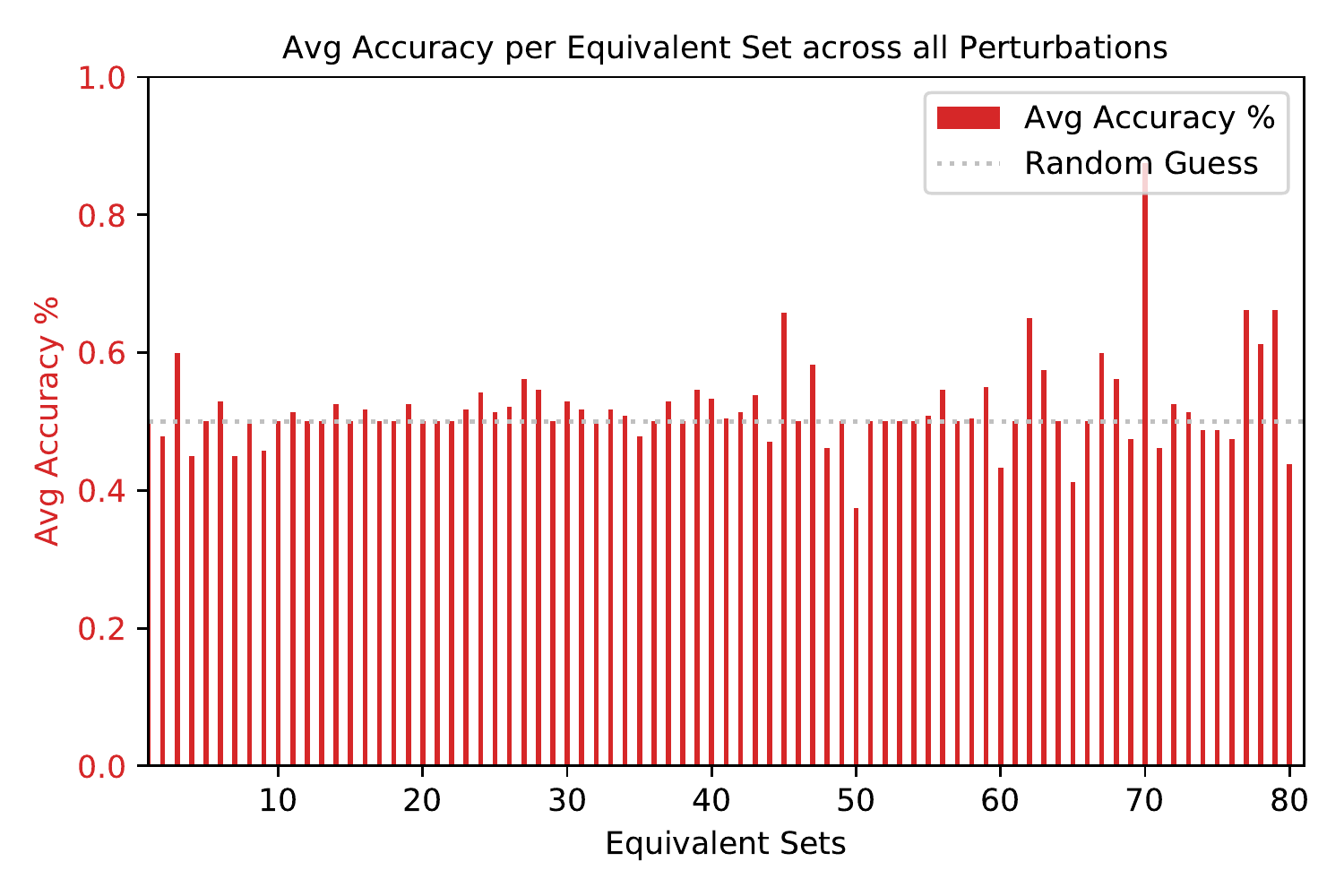}
	\caption{Results of average accuracy of RoBERTa-large on MWP. We can see that the PTLM makes random-guessing like predictions across all sets.}
	\label{fig:avg_per_set}
\end{figure}

\begin{table*}[!ht]
\resizebox{\textwidth}{!}{%
\begin{tabular}{lll}
\toprule
       linguistic perturbation & asymmetric perturbation &                                                                           probe \\
\midrule
                      original &                original &           A is wider than B, so A finds it harder to slip through cracks than B \\
                      original &      asymmetric\_premise &           B is wider than A, so A finds it easier to slip through cracks than B \\
                      original &   asymmetric\_conclusion &           A is wider than B, so B finds it easier to slip through cracks than A \\
                      negation &                original &   A is wider than B, so A does not find it easier to slip through cracks than B \\
                      negation &      asymmetric\_premise &   B is wider than A, so A does not find it harder to slip through cracks than B \\
                      negation &   asymmetric\_conclusion &   A is wider than B, so B does not find it harder to slip through cracks than A \\
                       antonym &                original &          A is wider than B, so A finds it easier to be blocked by cracks than B \\
                       antonym &      asymmetric\_premise &          B is wider than A, so A finds it harder to be blocked by cracks than B \\
                       antonym &   asymmetric\_conclusion &          A is wider than B, so B finds it harder to be blocked by cracks than A \\
                    paraphrase &                original &                A is wider than B, so A is worse at fitting into openings than B \\
                    paraphrase &      asymmetric\_premise &               B is wider than A, so A is better at fitting into openings than B \\
                    paraphrase &   asymmetric\_conclusion &               A is wider than B, so B is better at fitting into openings than A \\
          paraphrase\_inversion &                original &                A is wider than B, so A is more impeded by small openings than B \\
          paraphrase\_inversion &      asymmetric\_premise &                B is wider than A, so A is less impeded by small openings than B \\
          paraphrase\_inversion &   asymmetric\_conclusion &                A is wider than B, so B is less impeded by small openings than A \\
              negation\_antonym &                original &  A is wider than B, so A does not find it harder to be blocked by cracks than B \\
              negation\_antonym &      asymmetric\_premise &  B is wider than A, so A does not find it easier to be blocked by cracks than B \\
              negation\_antonym &   asymmetric\_conclusion &  A is wider than B, so B does not find it easier to be blocked by cracks than A \\
           negation\_paraphrase &                original &           A is wider than B, so A is not better at fitting into openings than B \\
           negation\_paraphrase &      asymmetric\_premise &            B is wider than A, so A is not worse at fitting into openings than B \\
           negation\_paraphrase &   asymmetric\_conclusion &            A is wider than B, so B is not worse at fitting into openings than A \\
 negation\_paraphrase\_inversion &                original &            A is wider than B, so A is not less impeded by small openings than B \\
 negation\_paraphrase\_inversion &      asymmetric\_premise &            B is wider than A, so A is not more impeded by small openings than B \\
 negation\_paraphrase\_inversion &   asymmetric\_conclusion &            A is wider than B, so B is not more impeded by small openings than A \\
\bottomrule
\end{tabular}
}
\caption{An example probe set––24 logically equivalent, but semantically different statements.}
\label{tab:perturbations}
\end{table*}

\clearpage
\begin{figure*}[!h]
	\centering
	\includegraphics[width=\textwidth]{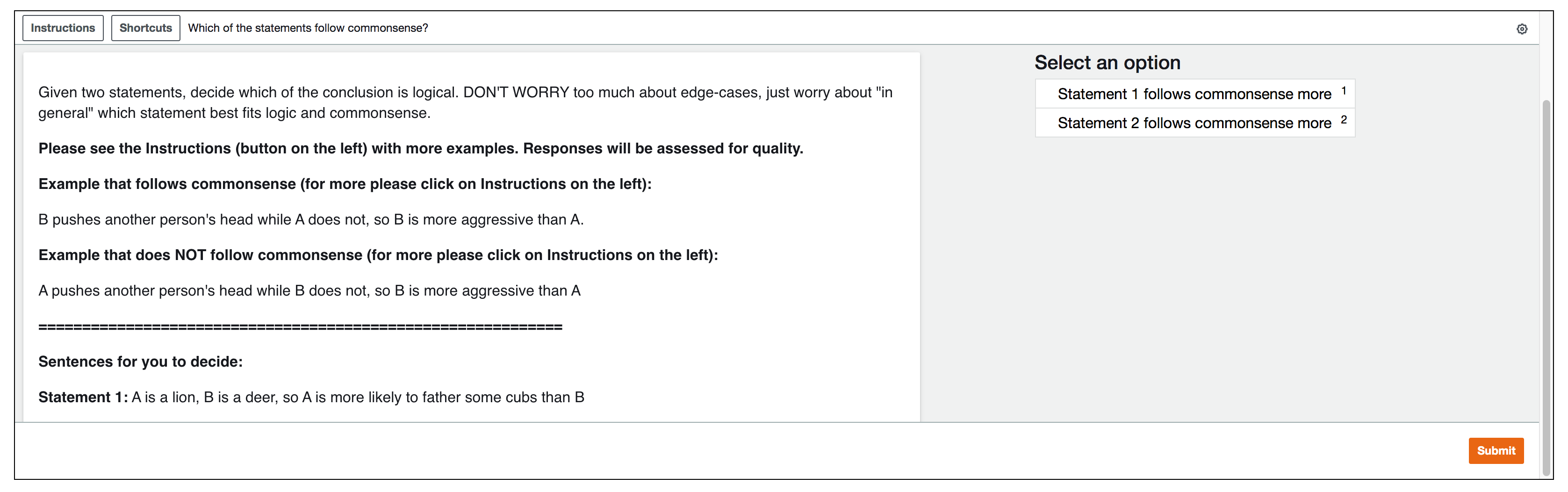}
	\caption{AMT annotation user interface for human verification on collected set.}
	\label{fig:turking}
\end{figure*}


\begin{table*}
\resizebox{\textwidth}{!}{%
\begin{tabular}{ll}
\toprule
template &                                                                                                  probe \\
\midrule
      1 &                     A is made out of glass and B is made out of stone, so A is more transparent than B \\
      1 &                          A is made out of cotton and B is made out of glass, so A is less sharp than B \\
      1 &                 A is made out of concrete and B is made out of paper, so A should be more heavy than B \\
      1 &                     A is made out of metal and B is made out of rubber, so A should float worse than B \\
      1 &                        A is made out of glass and B is made out of copper, so A is more fragile than B \\
      1 &                             A is made out of steel and B is made out of wool, so A is less soft than B \\
      1 &                      A is made out of wood and B is made out of glass, so A is more combustible than B \\
      1 &          A is made out of sponge and B is made out of nylon, so A is worse for water resistance than B \\
      1 &                     A is made out of copper and B is made out of concrete, so A is more ductile than B \\
      1 &                        A is made out of metal and B is made out of cloth, so A is less foldable than B \\
      1 &            A is made out of chocolate and B is made out of metal, so A is harder to keep frozen than B \\
      1 &         A is made out of metal and B is made out of dirt, so A is a better electrical conductor than B \\
      1 &               A is made out of stone and B is made out of helium, so A has a harder time flying than B \\
      1 &                         A is made out of honey and B is made out of water, so A is more viscous than B \\
      1 &                     A is made out of titanium and B is made out of rubber, so A is less elastic than B \\
      1 &                 A is made out of water and B is made out of methane, so A is more safe to store than B \\
      1 &  A is made out of mercury and B is made out of oxygen, so A is worse for your health to consume than B \\
      1 &        A is made out of wood and B is made out of fur, so A will more easily expand when heated than B \\
      1 &                    A is made out of concrete and B is made out of wood, so A is less penetrable than B \\
      1 &                 A is made out of glass and B is made out of tar, so A will reflect light better than B \\
      3 &                                 A makes the varsity team while B does not, so A is more skilled than B \\
      3 &             A is going to perform for people while B is not, so A finds it harder to be relaxed than B \\
      3 &                         A won the competition while B did not, so A finds it easier to be happy than B \\
      4 &                     A is able to concentrate more than B, so A finds it easier to be productive than B \\
      3 &                                            A bullies people while B does not, so A is less kind than B \\
      2 &                                                       A is B's boss, so A commands more respect than B \\
      4 &                                      A has more work than B, so A finds it harder to be at ease than B \\
      2 &                                        A has a crush on B, so A finds it harder to be relaxed around B \\
      4 &                              A has more dedication than B, so A will have a harder time failing than B \\
      2 &                                                   A is B's parent, so A initially takes more care of B \\
      2 &                                                             A is B's doctor, so A takes more care of B \\
      2 &                                              A hurt B's feelings, so A must be more insensitive than B \\
      2 &                                                  A is B's priest, so A spends less time sinning than B \\
      2 &                                               A is B's lawyer, so A is less ignorant of the law than B \\
      4 &                                  A has a lot less money than B, so A is less financially secure than B \\
      4 &    A watches more tv shows than B, so A is more capable of understanding pop-culture references than B \\
      2 &                          A always loses to B in tennis, so A is a less proficient tennis player than B \\
      2 &                                                A makes B late, so A has less reason to be annoyed at B \\
      4 &                                            A is a better friend than B, so A is more thoughtful than B \\
      2 &                                                  A is B's teacher, so A should be more informed than B \\
      4 &                                           A is smaller than B, so A is easier to put into a box than B \\
      4 &                                                  A is heavier than B, so A is better at sinking than B \\
      4 &                                  A is denser than B, so A should withstand piercing more easily than B \\
      4 &                                  A is wider than B, so A finds it harder to slip through cracks than B \\
      4 &                                               A is hotter than B, so A should be easier to melt than B \\
      4 &                                             A is more elastic than B, so A should bounce better than B \\
      4 &                                                A is tougher than B, so A is harder to rip apart than B \\
      4 &                                                    A is harder than B, so A is less comfortable than B \\
      4 &                                        A is taller than B, so A will cast a more lengthy shadow than B \\
      4 &                                     A is lighter than B, so A finds it harder to support weight than B \\
      4 &                        A has less momentum than B, so A has a worse ability to damage on impact than B \\
      4 &                                    A is more luminous than B, so A is more dangerous to look at than B \\
      4 &                                    A is more soluble than B, so A is harder to discern in water than B \\
      4 &                                              A is more pungent than B, so A is easier to detect than B \\
      4 &                           A is smaller than B, so A finds it harder to displace liquid in a tub than B \\
      4 &                              A is shorter than B, so A is worse for keeping things out of reach than B \\
      4 &                                             A is larger than B, so A is more difficult to carry than B \\
      4 &                           A is more taut than B, so A is worse at withstanding additional force than B \\
      4 &                                 A is much hotter than B, so A will be more painful to hold onto than B \\
      4 &                       A is more magnetic than B, so A is harder to separate from another magnet than B \\
\bottomrule
\end{tabular}
}
\caption{Sixty probes and their corresponding logical templates}
\label{tab:allprobes}
\end{table*}

\newpage
\begin{table*}[]
\resizebox{\textwidth}{!}{%
\begin{tabular}{@{}cc|ccc|ccc@{}}
\toprule
\multicolumn{1}{l}{}                                & \multicolumn{1}{l|}{}  & \multicolumn{3}{c|}{BERT-base}  & \multicolumn{3}{c}{BERT-large} \\
\multicolumn{1}{l}{}                                & \multicolumn{1}{l|}{}  & Easy Set & Hard Set & Joint Set & Easy Set & Hard Set & Joint Set \\ \midrule
\multicolumn{2}{c|}{Zero-shot}                                                & 49.32    & 49.7     & 49.56     & 49.15    & 49.35    & 49.27     \\ \midrule
\multicolumn{1}{c|}{\multirow{4}{*}{Low resource}}  & 10\%                   & 56.38    & 49.85    & 52.37     & 63.08    & 50.18    & 55.16     \\
\multicolumn{1}{c|}{}                               & 20\%                   & 59.3     & 50.5     & 53.89     & 65.64    & 50.37    & 56.26     \\
\multicolumn{1}{c|}{}                               & 30\%                   & 55.09    & 50.22    & 52.1      & 65.3     & 50.27    & 56.07     \\
\multicolumn{1}{c|}{}                               & 50\%                   & 60.62    & 50.33    & 54.3      & 63.48    & 50.25    & 55.35     \\ \midrule
\multicolumn{1}{c|}{\multirow{3}{*}{Full resource}} & with 1 novel entity    & 54.26    & 50.5     & 51.95     & 61.02    & 49.46    & 53.97     \\
\multicolumn{1}{c|}{}                               & with 5 novel entities  & 64.48    & 50.58    & 55.94     & 69.92    & 49.72    & 57.51     \\
\multicolumn{1}{c|}{}                               & with 10 novel entities & 82.58    & 51.18    & 63.3      & 85.93    & 50.64    & 64.74     \\ \midrule
\multicolumn{2}{c|}{100k}                                                     & 45.4     & 50.09    & 46.03     & 46.14    & 50.93    & 47.02     \\ \bottomrule
\end{tabular}
}
\end{table*}

\begin{table*}[]
\resizebox{\textwidth}{!}{%
\begin{tabular}{@{}cc|ccc|ccc@{}}
\toprule
\multicolumn{1}{l}{}                                & \multicolumn{1}{l|}{}  & \multicolumn{3}{c|}{RoBERTa-base} & \multicolumn{3}{c}{RoBERTa-large} \\
\multicolumn{1}{l}{}                                & \multicolumn{1}{l|}{}  & Easy Set  & Hard Set  & Joint Set & Easy Set  & Hard Set  & Joint Set  \\ \midrule
\multicolumn{2}{c|}{Zero-shot}                                               & 53.6      & 49.39     & 55.32     & 51.81     & 49.69     & 58.79      \\ \midrule
\multicolumn{1}{c|}{\multirow{4}{*}{Low resource}}  & 10\%                   & 59.08     & 49.36     & 53.1      & 59.93     & 50.65     & 54.23      \\
\multicolumn{1}{c|}{}                               & 20\%                   & 60.53     & 49.41     & 53.7      & 64.31     & 50.31     & 55.71      \\
\multicolumn{1}{c|}{}                               & 30\%                   & 60.79     & 49.93     & 54.21     & 65.47     & 50.95     & 56.55      \\
\multicolumn{1}{c|}{}                               & 50\%                   & 64.11     & 49.4      & 55.07     & 75.88     & 52.58     & 61.57      \\ \midrule
\multicolumn{1}{c|}{\multirow{3}{*}{Full resource}} & with 1 novel entity    & 64.24     & 49.89     & 55.43     & 79.1      & 52.84     & 62.97      \\
\multicolumn{1}{c|}{}                               & with 5 novel entities  & 82.32     & 49.93     & 62.43     & 87.03     & 51.69     & 65.32      \\
\multicolumn{1}{c|}{}                               & with 10 novel entities & 85.63     & 50.64     & 64.14     & 84.74     & 51.08     & 64.07      \\ \midrule
\multicolumn{2}{c|}{100k}                                                    & 72.35     & 50.02     & 70.06     & 78.13     & 53.92     & 73.71      \\ \bottomrule
\end{tabular}
}
\end{table*}

\begin{table*}[]

\begin{tabular}{@{}cc|ccc@{}}
\toprule
\multicolumn{1}{l}{}                                & \multicolumn{1}{l|}{}  & \multicolumn{3}{c}{GPT2}                            \\
\multicolumn{1}{l}{}                                & \multicolumn{1}{l|}{}  & Easy Set & Hard Set & \multicolumn{1}{c}{Joint Set} \\ \midrule
\multicolumn{2}{c|}{Zero-shot}                                               & 51.27    & 49.6     & 50.1                           \\ \midrule
\multicolumn{1}{c|}{\multirow{4}{*}{Low resource}}  & 10\%                   & 50.57    & 49.91    & 50.29                          \\
\multicolumn{1}{c|}{}                               & 20\%                   & 48.22    & 49.33    & 49.01                          \\
\multicolumn{1}{c|}{}                               & 30\%                   & 48.18    & 49.2     & 48.97                          \\
\multicolumn{1}{c|}{}                               & 50\%                   & 50.44    & 49.87    & 49.96                          \\ \midrule
\multicolumn{1}{c|}{\multirow{3}{*}{Full resource}} & with 1 novel entity    & 55.95    & 49.34    & 52.16                          \\
\multicolumn{1}{c|}{}                               & with 5 novel entities  & 66.3     & 49.53    & 55.98                          \\
\multicolumn{1}{c|}{}                               & with 10 novel entities & 71.6     & 49.91    & 58.25                          \\ \midrule
\multicolumn{2}{c|}{100k}                                                    & 32.94    & 49.16    & 35.21                          \\ \bottomrule
\end{tabular}
\end{table*}

\begin{table*}[]
\begin{tabular}{@{}cc|cc@{}}
\toprule
                                                    &                        & \multicolumn{2}{c}{BART} \\
                                                    &                        & EASY        & HARD        \\ \midrule
\multicolumn{2}{c|}{Zeroshot}                                                & 47.46       & 50.37       \\ \midrule
\multicolumn{1}{c|}{\multirow{4}{*}{Low resource}}  & 10\%                   & 60.55       & 49.72       \\
\multicolumn{1}{c|}{}                               & 20\%                   & 61.58       & 49.91       \\
\multicolumn{1}{c|}{}                               & 30\%                   & 63.54       & 49.96       \\
\multicolumn{1}{c|}{}                               & 50\%                   & 61.65       & 49.68       \\ \midrule
\multicolumn{1}{c|}{\multirow{3}{*}{Full resource}} & with 1 novel entity    & 63.94       & 49.31       \\
\multicolumn{1}{c|}{}                               & with 5 novel entities  & 69.05       & 49.73       \\
\multicolumn{1}{c|}{}                               & with 10 novel entities & 79.73       & 50.52       \\ \midrule
\multicolumn{2}{c|}{100k}                                                    & 50.87       & 49.95       \\ \bottomrule
\end{tabular}
\end{table*}

\begin{table*}[]
\begin{tabular}{@{}cc|cc@{}}
\toprule
                                   &                        & \multicolumn{2}{c}{ERNIE} \\
                                   &                        & EASY         & HARD        \\ \midrule
\multicolumn{2}{c|}{Zeroshot}                               & 47.97        & 50.31       \\ \midrule
\multicolumn{1}{c|}{\multirow{3}{*}{Low resource}}  & 1\%                    & 63.11        & 50.22       \\
\multicolumn{1}{c|}{}              & 3\%                    & 72.91        & 48.89       \\
\multicolumn{1}{c|}{}              & 5\%                    & 87.42        & 49.70       \\ \midrule
\multicolumn{1}{c|}{\multirow{3}{*}{Full resource}} & with 1 novel entity    & 87.66        & 50.06       \\
\multicolumn{1}{c|}{}              & with 5 novel entities  & 87.65        & 50.03       \\
\multicolumn{1}{c|}{}              & with 10 novel entities & 87.36        & 49.77       \\ \midrule
\multicolumn{2}{c|}{100k}                                   & 69.25        & 50.95       \\ \bottomrule
\end{tabular}
\end{table*}

\end{document}